\documentclass[11pt, a4paper, confidential, copyright, goog]{google}

\usepackage[authoryear, sort&compress, round]{natbib}
\usepackage{hyperref}
\usepackage{url}
\usepackage{graphicx}
\usepackage{booktabs}
\usepackage{xcolor}
\usepackage{pifont}
\usepackage{wrapfig}  
\usepackage{xspace}
\usepackage{subcaption}
\usepackage[table]{xcolor}
\usepackage[most]{tcolorbox}
\usepackage{multirow} 
\usepackage{algorithm}
\usepackage{algpseudocode}

\usepackage{geometry}
\usepackage{caption}          % for \captionsetup
\usepackage{listings}         % code listings
\usepackage{upquote}          % makes quotes upright in tt fonts
\usepackage[most]{tcolorbox}  % tcolorbox base
\tcbuselibrary{listings,breakable,skins} % tcolorbox libs for listings
\lstdefinestyle{boxedcode}{
  basicstyle=\ttfamily\small, 
  columns=fullflexible,
  keepspaces=true,
  breaklines=true,
  upquote=true,
  showstringspaces=false
}

\uselogo{} 

\title{Budget-Aware Tool Use Enables Effective Agent Scaling}
\correspondingauthor{tengxiao@ucsb.edu, \{zifengw, chenyulee\}@google.com}

\author[1*]{Tengxiao Liu}
\author[2$\dagger$]{Zifeng Wang}
\author[3]{Jin Miao}
\author[2]{I-Hung Hsu}
\author[2]{Jun Yan}
\author[2]{Jiefeng Chen}
\author[2]{Rujun Han}
\author[4]{Fangyuan Xu}
\author[2]{Yanfei Chen}
\author[2]{Ke Jiang}
\author[2]{Samira Daruki}
\author[3]{Yi Liang}
\author[1]{William Yang Wang}
\author[2]{Tomas Pfister}
\author[2$\dagger$]{Chen-Yu Lee}

\affil[1]{UC Santa Barbara}
\affil[2]{Google Cloud AI Research}
\affil[3]{Google DeepMind}
\affil[4]{New York University}

\begin{abstract}
Scaling test-time computation has been extended from language model reasoning to tool-augmented agents, where scaling involves not only thinking in tokens but also acting via tool calls that directly constrain environmental interaction.
However, we found that simply increasing the tool-call budget fails to improve performance, as agents lack ``budget awareness'' and quickly hit a performance ceiling.
We study how to scale such agents effectively under explicit tool-call budgets, focusing on web search agents. 
We first introduce the \textbf{Budget Tracker}, a lightweight plug-in that provides the agent with continuous budget awareness, enabling simple yet effective scaling. 
We further develop \textbf{BATS} (\textbf{B}udget \textbf{A}ware \textbf{T}est-time \textbf{S}caling), an advanced framework that leverages this awareness to dynamically adapt its planning and verification strategy.
To analyze cost-performance scaling in a controlled manner, we formalize a unified cost metric that jointly accounts for token and tool consumption. We provide the first systematic study on budget-constrained agents, showing that budget-aware methods produce more favorable scaling curves and push the cost-performance Pareto frontier.
Our work offers empirical insights toward a more transparent and principled understanding of scaling in tool-augmented agents.
Our code is available at \url{https://github.com/google-research/budget-aware-agent}.
\end{abstract}

\newcommand{\cmark}{\ding{51}}
\newcommand{\xmark}{\ding{55}}

\newcommand{\method}{BATS\xspace}

\begin{document}

\maketitle

\section{Introduction}

Scaling test-time compute in large language models (LLMs) helps improve the performance across a wide range of reasoning tasks~\citep{DBLP:conf/iclr/WuSLWY25,DBLP:journals/corr/abs-2408-03314, DBLP:journals/corr/abs-2501-19393, DBLP:journals/corr/abs-2501-19306}. 
Mainstream strategies such as sequential~\citep{DBLP:conf/nips/MadaanTGHGW0DPY23} and parallel scaling~\citep{DBLP:conf/iclr/0002WSLCNCZ23,DBLP:journals/corr/abs-2407-21787} increase effective computation, enabling deeper reflection and output refinement with substantial gains in answer quality~\citep{DBLP:journals/corr/abs-2503-24235}. These successes motivate recent efforts to extend test-time scaling to tool-augmented agents~\citep{DBLP:journals/corr/abs-2506-12928, DBLP:journals/corr/abs-2508-00890}, where LLMs are equipped with various tools to interact with the external environment.

Test-time scaling for tool-augmented agents expands both thinking (tokens) and acting (tool calls). While textual reasoning tasks are usually scaled by thinking with more tokens, solely increasing the internal thinking effort in agents proves to be suboptimal and does not always translate to performance scaling~\citep{DBLP:journals/corr/abs-2506-07976}.
Unlike controlling the token count in textual reasoning~\citep{DBLP:conf/acl/HanWFZM025,DBLP:journals/corr/abs-2504-13367}, in agent tasks such as web browsing, the number of tool calls directly determines the depth and breadth of exploration~\citep{anthropic2025context, miromindteam2025mirothinkerpushingperformanceboundaries}, defining the effective boundary of external information access.

However, simply granting agents more tool-call resources does not guarantee better performance. Our analysis reveals a critical bottleneck: \emph{standard agents lack inherent budget awareness}. Without explicit signals, they often perform shallow searches~\citep{lu2025deepdiveadvancingdeepsearch} and fail to utilize additional resources even when available. As our empirical findings later demonstrate, standard agents quickly hit a ``performance ceiling'', saturating their capabilities regardless of extra budget allocated.
The core challenge is not spending more, but managing diminishing and uncertain marginal gains from additional tool calls.
% The challenge is not simply spending more, but spending wisely: the marginal benefit per tool call is uncertain, so every call must be spent strategically.

\begin{figure}[t] 
    % \vspace{-10pt}
    \centering
    \includegraphics[width=0.8\linewidth]{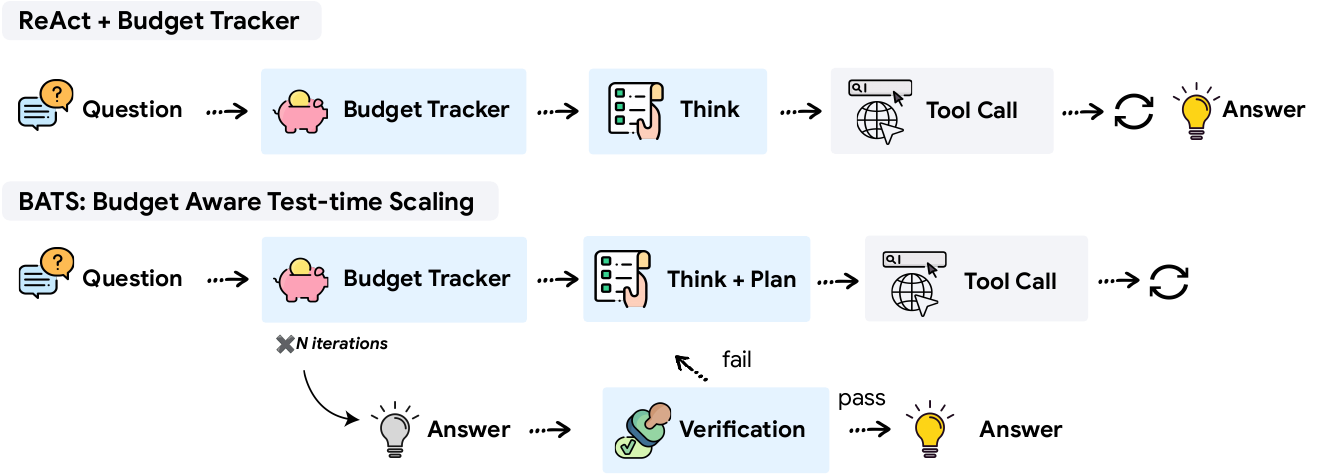} 
    \caption{Budget Tracker is a lightweight plug-in that can be applied to both a standard ReAct agent (top) and more advanced orchestration frameworks like BATS (bottom). In this figure, blue boxes highlight modules that adapt to the budget.}
    \label{fig:bt_bats}
    % \vspace{-10pt}
\end{figure}

This unique complexity brings up a critical research question: How can tool-augmented agents scale \emph{effectively} by making the best use of a given resource budget?
We study this question in a budget-constrained setting, grounding our analysis in search agents equipped with search and browse tools, which are widely used in practice~\citep{google2025geminideepresearch, openai2025deepresearch} and inherently require extensive tool calls to collect external information.
To ensure a fair and transparent comparison, we formalize a unified cost metric that jointly accounts for the economic costs of both internal token consumption and external tool interactions. This metric enables us to trace a true cost-performance scaling trend.

We first develop an intuitive, light-weight approach: \textbf{Budget Tracker} (Figure~\ref{fig:bt_bats} (top)), which is a plug-and-play module compatible with ReAct-based agents~\citep{DBLP:conf/iclr/YaoZYDSN023}. Budget Tracker provides the agent with a continuous signal of resource availability, proving to be a simple yet effective method. Through extensive experiments, we show that this simple plug-in improves performance across various budget constraints. Combining this explicit budget awareness with traditional test-time scaling strategies consistently enables more effective scaling behaviors while continuously pushing the cost-performance Pareto frontier.

% While simple awareness is effective, it still operates within the agent's predefined logic. 
To further break the scaling ceiling, we introduce \textbf{\method} (\textbf{B}udget \textbf{A}ware \textbf{T}est-time \textbf{S}caling) (Figure~\ref{fig:bt_bats} (bottom)), a framework that fully internalizes resource constraints to maximize agent performance under any given budget.
\method maintains a continuous signal of remaining resources and uses it to dynamically adapt its behavior: a planning module adjusts stepwise effort to match the current budget, while a verification module decides whether to dig deeper into a promising lead or pivot to alternative paths.
% To further break the scaling ceiling and fully internalize resource constraints, we introduce \textbf{\method} (\textbf{B}udget \textbf{A}ware \textbf{T}est-time \textbf{S}caling) (Figure~\ref{fig:bt_bats} (bottom)), a framework designed to maximize agent performance under any given budget. 
% \method maintains a continuous signal of remaining resources, using this information to dynamically adapt its behavior. A planning module adjusts stepwise effort to match the current budget, while a verification module decides whether to ``dig deeper'' into a promising lead or ``pivot'' to alternative paths based on resource availability.

Systematic experiments under varying budgets show that \method achieves higher performance with fewer tool calls and lower overall cost than competing methods, producing more favorable scaling curves and better cost-performance trade-offs. These findings highlight the importance of explicitly accounting for cost in agent test-time scaling, enabling agents that are both effective and efficient.
% These findings demonstrate the potential of budget-aware design for creating effective and efficient tool-augmented agents, highlighting the importance of explicitly accounting for cost in agent test-time scaling.
% Our empirical study systematically evaluates the relationship between overall resource cost and task performance by comparing different scaling frameworks under varying budgets.
% The results show that \method produces more favorable scaling curves: it achieves higher performance while using fewer tool calls and incurring lower overall cost than competing methods.

We summarize our contributions as follows:
\begin{itemize}
    \item[\textbf{(1)}] We provide the first systematic study of budget-constrained tool-use agents, formalizing agent test-time scaling with explicit tool-call budgets and introducing a unified cost metric.
    \item[\textbf{(2)}] We introduce \textbf{Budget Tracker}, a lightweight plug-in compatible with any agent orchestration framework that enables effective budget-aware tool use.
    \item[\textbf{(3)}] We develop \textbf{\method}, a budget-aware framework that dynamically adapts planning and verification strategies based on real-time resource tracking, flexibly switching between deepening a lead and branching to alternatives.
    \item[\textbf{(4)}] We conduct systematic experiments under varying budgets and unified costs with search agents, demonstrating that \method is more cost-effective and yields more favorable scaling curves and better cost–performance trade-offs.
\end{itemize}

\section{Problem Formulation}

\subsection{Agent Test-time Scaling}

We formulate agent test-time scaling as how an agent's performance scales with its budget for external tool-call interaction, refining the broader concept of test-time interaction scaling in~\citet{DBLP:journals/corr/abs-2506-07976}. To make agent test-time scaling cost-effective, an ideal agent should be able to achieve its best possible performance under an arbitrary budget constraint on the scaling curve. To this end, our target is to design an effective and efficient agent framework, $\pi$, that maximizes answer accuracy while adhering to a strict tool-call budget.

Assume the agent is equipped with a set of $K$ tools as $\mathcal{T}=\{t_1, \dots, t_{K}\}$. For question $x \in \mathcal{X}$, the agent works under a budget $\mathbf{b}=(b_1,\dots,b_K)$, where $b_i$ is the max number of invocations of tool $t_i \in \mathcal{T}$.
Let $\hat{y}_\pi(x)$ denote the agent's predicted answer for question $x$ with ground truth $y(x)$ and let $c_i(x;\pi)$ be the realized number of calls to tool $t_i$ on $x$.
We formulate the agent test-time scaling problem as a budget-constrained optimization objective:

\begin{equation}
\label{eq:tts-per-tool-acc}
\begin{aligned}
\max_{\pi}\quad
& \mathrm{Acc}_{\mathbf{b}}(\pi)
\;\;=\;\; \mathbb{E}_{x}\big[\,\mathbf{1}\{\hat{y}_\pi(x)=y(x)\}\,\big] \\[2pt]
\text{s.t.}\quad
& c_i(x;\pi) \;\le\; b_i,\ 
\text{for all } i=1,\dots,K,\ \text{and every } x \in \mathcal{X}
\end{aligned}
\end{equation}

Here the objective is the expected accuracy over all questions. The constraint ensures that for any given question, the number of realized calls for each tool never exceeds its allocated budget. By evaluating the agent performance at various budget levels, we can trace the performance-cost curve, which characterizes the agent's test-time scaling behavior, showing how effectively it leverages budget resources to improve its problem solving capabilities.

\textbf{Budget vs. Cost.} We distinguish between the preset budget constraint, which specifies the maximum number of tool calls available to the agent, and the realized cost, which reflects the resources actually consumed during execution. While the budget imposes a hard upper limit, the final cost depends on the agent's strategy. For consistent comparison, we introduce in Section~\ref{sec:prob_instantiation} a unified post-hoc cost metric for analyzing agents' test-time scaling.

% \textbf{Choice of Budget.} Among possible constraints, we prioritize a tool-call budget over a token-based budget for its \textit{relevance}, \textit{consistency}, and \textit{practicability}. A tool-call limit offers a more relevant and direct constraint on an agent's ability to acquire external knowledge than the tokens used for internal reasoning. This choice is also consistent with and justified by established practices~\citep{DBLP:journals/corr/abs-2506-07976, miromindteam2025mirothinkerpushingperformanceboundaries}. Furthermore, it offers greater practicability, as it is often non-trivial to pre-determine an appropriate token budget for complex, multi-step agentic tasks. While the budget is defined by tool calls, we still incorporate token usage into our unified cost metric (Section~\ref{sec:prob_instantiation}) to ensure a more fair and transparent comparison.

\textbf{Choice of Budget.} We prioritize a tool-call budget over a token-based budget for its \textit{relevance}: tool calls more directly constrain an agent's ability to acquire external knowledge; \textit{consistency}: this choice aligns with established practices~\citep{DBLP:journals/corr/abs-2506-07976, miromindteam2025mirothinkerpushingperformanceboundaries}; and \textit{practicability}: pre-determining an accurate token budget for multi-step agentic tasks is often impractical. While the budget is defined by tool calls, we still incorporate token usage into our unified cost metric (Section~\ref{sec:prob_instantiation}) to ensure a more fair and transparent comparison.

\subsection{Problem Instantiation with Search Agent}
\label{sec:prob_instantiation}
We instantiate test-time scaling through search agents: LLMs that reason over retrieved evidence to answer information-seeking questions $x$. This setting offers broad applicability and established benchmarks. Using a widely adopted ReAct-style loop~\citep{DBLP:conf/iclr/YaoZYDSN023}, the agent alternates between reasoning and action using two tools:

\begin{itemize}
    \item[\textbf{(1)}] \textbf{Search.} This tool helps perform a standard search engine query. Given a text query, it returns a list of search results, each including a title, a brief snippet, and a URL.
    \item[\textbf{(2)}] \textbf{Browse.} Given a specific URL, this tool scrapes the full content of the corresponding webpage, providing detailed information that is often unavailable in a search snippet.
\end{itemize}

% In our work, we instantiate the test-time scaling problem with search agent, a setting selected for its broad applicability and the presence of established benchmarks.  A search agent is an LLM that answers an information-seeking question $x$ by retrieving external evidence and reasoning over it. The agent follows an iterative ReAct-style loop~\citep{DBLP:conf/iclr/YaoZYDSN023}, alternating between internal thinking and external actions. The agent has access to two primary tools for interacting with the world:

\textbf{Unified Cost Metric.} We define total cost as the sum of token and tool call usage, mapped to their corresponding economic costs for a unified metric.

\begin{itemize}
    \item[\textbf{(1)}] \textbf{Token cost.} This represents the agent's internal cognitive effort, including its reasoning, planning, and parametric knowledge processing. Token costs are calculated based on the pricing of the model provider, distinguishing among input, output, and cache hit tokens. In multi-round iterative frameworks, the output of iteration $i$ becomes part of the input for iteration $i+1$. Any overlap is a cache hit, thereby lowering the token consumption cost.
    \item[\textbf{(2)}] \textbf{Tool call cost.} This captures the agent's active interaction with the environment through information-seeking actions. Each invocation (e.g., a search query or browsing request) incurs a cost determined by the corresponding API provider's pricing.
\end{itemize}

% This represents the agent's active interaction with the environment through information-seeking actions. Each invocation of an external service (e.g., a search query or a browsing request) incurs a cost determined by the pricing of the corresponding API or third-party provider. 

The total unified cost, $C_{\textit{unified}}(x;\pi)$, for solving a given question $x$ under policy $\pi$ is the sum of the token cost and the tool call cost. Let $c_i(x;\pi)$ be the number of actual invocations to tool $t_i$, and $P_i$ be the economic cost per invocation of $t_i$. We define:
\begin{equation}
\label{eq:unified-cost-definition}
C_{\textit{unified}}(x;\pi) = \underbrace{c_{\textit{token}}(x;\pi)}_{\text{Token Cost}} + \underbrace{\sum_{i=1}^{K} c_i(x;\pi) \cdot P_i}_{\text{Total Tool Cost}}
\end{equation}
Here, $c_{\textit{token}}(x;\pi)$ represents the total cost incurred from token consumption during the reasoning process for question $x$. This formulation allows us to uniformly analyze the actual incurred costs. These two dimensions are inherently coupled: additional tool calls generally increase token consumption, as the agent must process and reason over the retrieved external information. Measuring this unified cost under varying budget constraints $\mathbf{b}$ enables a more comprehensive and consistent analysis across different policies.

\section{Budget Awareness}

\begin{figure*}[h] 
    \centering
    \includegraphics[width=0.8\textwidth]{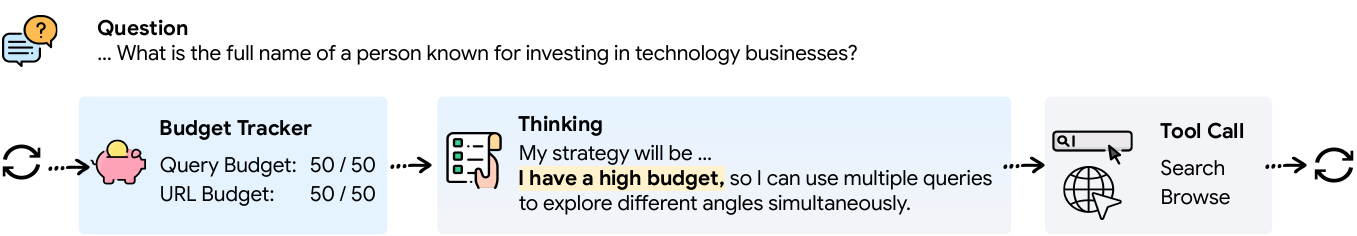} 
    \caption{At each interaction round, the agent is provided with its current and remaining budget through the budget tracker before generating the next thinking and tool calls.}
    \label{fig:budget_tracker}
\end{figure*}

\subsection{Budget Tracker}

Each tool invocation introduces non-negligible costs in context and computation. We hypothesize that providing explicit budget signals enables the model to internalize resource constraints and adapt its strategy without additional training.

We introduce \textbf{Budget Tracker}, a lightweight, plug-and-play module that surfaces real-time budget states within the agent’s reasoning loop.
At the first iteration, the tracker provides a brief policy guideline describing the budget regimes and corresponding tool-use recommendations~(see Appendix~\ref{app:prompts}).
At each subsequent iteration, it appends a budget status block showing the remaining and used budgets for each tool. Though simple, this mechanism allows the agent to condition its reasoning on the updated resource state:

\begin{tcolorbox}[
  enhanced,
  breakable,
  colback=white,
  colframe=black!10!white, 
  coltitle=black!80!white, 
  title={Budget Tracker},
  % fonttitle=\large\bfseries,
]
{
\small
<budget> \\
Tool$_{i}$ Budget Used: \#\#, Tool$_{i}$ Budget Remaining: \#\# \\
Make the best use of the available resources. \\
</budget>
}
\end{tcolorbox}

% Though simple, Budget Tracker operates at the prompt level to provide resource availability, allowing the agent to condition its reasoning steps on the updated resource state.

% Despite its simplicity, Budget Tracker operates purely at the prompt level and makes the agent explicitly aware of its resource consumption and remaining budget, enabling it to condition subsequent reasoning steps on the updated resource state.

\subsection{Scaling with Budget Awareness} 
To better understand how agent behaviors scale under different budget constraints, we study two mainstream test-time scaling paradigms: \emph{sequential scaling} and \emph{parallel scaling}. 
In \textit{sequential scaling}, we adopt the budget-forcing strategy~\citep{DBLP:journals/corr/abs-2501-19393} by appending the message when the agent proposes an answer: ``Wait, you still have remaining tool budget, use more search and browse tools to explore different information sources before concluding.''
This encourages the model to reevaluate its reasoning and make better use of its available tool calls.
In \textit{parallel scaling}, we fix the per-run budget and conduct multiple independent runs, reporting Majority Vote, Best-of-N and Pass@N results following standard test-time scaling practice~\citep{DBLP:journals/corr/abs-2504-12516}. See Appendix~\ref{app:pass@n} for more details.

\subsection{Results and Analysis}

% We evaluate Budget Tracker mainly on three information-seeking QA datasets requiring external search: BrowseComp~\citep{DBLP:journals/corr/abs-2504-12516}, BrowseComp-ZH~\citep{DBLP:journals/corr/abs-2504-19314} and HLE-Search~\citep{DBLP:journals/corr/abs-2501-14249, DBLP:journals/corr/abs-2507-16075}. We also evaluate general conversational agents using $\tau^2$-bench~\citep{DBLP:journals/corr/abs-2506-07982} in Appendix~\ref{app:tau2}.
% Dataset details and full experimental setup are provided in Section~\ref{sec:exp}.
% Building on the ReAct framework, we incorporate the Budget Tracker immediately after each tool response to inform the agent of its remaining budget. Each tool is assigned a budget of 100, and the reasoning process stops when any budget is exhausted or the agent reaches a final answer. 

We evaluate Budget Tracker on three information-seeking benchmarks (BrowseComp~\citep{DBLP:journals/corr/abs-2504-12516}, BrowseComp-ZH~\citep{DBLP:journals/corr/abs-2504-19314} and HLE-Search~\citep{DBLP:journals/corr/abs-2501-14249, DBLP:journals/corr/abs-2507-16075}; details in Section~\ref{sec:exp}) and on conversational tasks ($\tau^2$-bench~\citep{DBLP:journals/corr/abs-2506-07982}; Appendix~\ref{app:tau2}).
Building on the ReAct framework, the Budget Tracker is inserted immediately after each tool response. Each tool is assigned a budget of 100, and the process stops when any budget is exhausted or the agent reaches a final answer.

% \begin{table}[ht]
% \centering
% % \small
% \caption{Results show consistent accuracy gains with Budget Tracker integration. The results are averaged over 3 independent runs. 
% % BC, BC-ZH, and HLE denote BrowseComp, BrowseComp-ZH, and the HLE-Search.
% }\label{tab:react_bab}
% \resizebox{0.8\linewidth}{!}{
% \begin{tabular}{llrrr}
% \toprule
% \textbf{Model} & \textbf{Method} & \textbf{BrowseComp} & \textbf{BrowseComp-ZH} & \textbf{HLE-Search} \\
% \midrule
% \multirow{2}{*}{Gemini-2.5-Pro} 
% & ReAct & 12.6{\scriptsize $\pm$1.2} & 31.5 & 20.5\\
% & + Budget Tracker & \textbf{14.6}{\scriptsize $\pm$1.2}  & \textbf{32.9} & \textbf{21.8}\\
% \midrule
% \multirow{2}{*}{Gemini-2.5-Flash} 
% & ReAct & 9.7 & 26.5 & 14.7\\
% & + Budget Tracker & \textbf{10.7}  & \textbf{28.7} & \textbf{17.3} \\
% \midrule
% \multirow{2}{*}{Claude-Sonnet-4} 
% & ReAct & 12.2 & 29.1 & 20.5 \\
% & + Budget Tracker & \textbf{14.0}  & \textbf{31.1} & \textbf{23.0} \\
% \bottomrule
% \end{tabular}
% }
% \vspace{-3pt}
% \end{table}

\begin{table}[ht]
\centering
% \small

% \resizebox{0.85\linewidth}{!}{
\begin{tabular}{llrrr}
\toprule
\textbf{Model} & \textbf{Method} & \textbf{BrowseComp} & \textbf{BrowseComp-ZH} & \textbf{HLE-Search} \\
\midrule
\multirow{2}{*}{Gemini-2.5-Pro} 
& ReAct & 12.6{\scriptsize $\pm$1.2} & 31.5{\scriptsize $\pm$1.5} & 20.5{\scriptsize $\pm$0.5}\\
& + Budget Tracker & \textbf{14.6}{\scriptsize $\pm$1.2}  & \textbf{32.9}{\scriptsize $\pm$0.5} & \textbf{21.8}{\scriptsize $\pm$1.0}\\
\midrule
\multirow{2}{*}{Gemini-2.5-Flash} 
& ReAct & 9.7{\scriptsize $\pm$0.5} & 26.5{\scriptsize $\pm$0.7} & 14.7{\scriptsize $\pm$1.1}\\
& + Budget Tracker & \textbf{10.7}{\scriptsize $\pm$0.2}  & \textbf{28.7}{\scriptsize $\pm$1.7} & \textbf{17.3}{\scriptsize $\pm$2.3} \\
\midrule
\multirow{2}{*}{Claude-Sonnet-4} 
& ReAct & 12.2{\scriptsize $\pm$0.6} & 29.1{\scriptsize $\pm$1.7} & 20.5{\scriptsize $\pm$0.5} \\
& + Budget Tracker & \textbf{14.0}{\scriptsize $\pm$0.5}  & \textbf{31.1}{\scriptsize $\pm$1.2} & \textbf{23.0}{\scriptsize $\pm$0.5} \\
\bottomrule
\end{tabular}
% }
\caption{Results show consistent accuracy gains with Budget Tracker integration across different models and benchmarks. The results are averaged over 3 independent runs. 
% BC, BC-ZH, and HLE denote BrowseComp, BrowseComp-ZH, and the HLE-Search.
}\label{tab:react_bab}
\end{table}

\textbf{Budget Tracker enables higher performance under the same budget.}
Table~\ref{tab:react_bab} compares ReAct with and without the Budget Tracker under identical budget limits. Across different models, adding the tracker consistently improves accuracy on all the datasets.  Making budget signals explicit encourages more strategic and positive agent tool-use behavior.

\begin{table}[h]
\centering
% \scalebox{0.85}{
\begin{tabular}{lrrrrr}
\toprule
\multirow{2}{*}{\textbf{Method}} & 
\multirow{2}{*}{\textbf{Budget}} & 
\multirow{2}{*}{\textbf{Acc (\%)}} &
\multicolumn{3}{c}{\textbf{Consumed resources}} \\ 
\cmidrule(lr){4-6}
 &  &  & \textbf{\# search} & \textbf{\# browse} & \textbf{cost (¢)} \\
\midrule
\multirow{3}{*}{ReAct} 
 & 10  & 10.3 & 7.87  & 0.60 & 5.2 \\
 & 30  & 10.5 & 13.77 & 1.33 & 8.0 \\
 & 100 & 12.6 & 14.24 & 1.36 & 9.9 \\
\midrule
+ Budget Tracker & 10 & \textbf{12.8} & 8.48 & 1.09 & 6.8 \\
\bottomrule
\end{tabular}
% }
\caption{Comparison under different budgets using Gemini-2.5-Pro. With 10$\times$ less budget (10 vs.\ 100), Budget Tracker (+ BT) achieves comparable accuracy while requiring fewer tool calls and lower total cost, highlighting the benefits of budget awareness.}
\label{tab:react_less_cost}

\end{table}

% \begin{figure}[h]
%     \centering
%     % \vspace{-10pt} 
%     \includegraphics[width=0.98\linewidth]{assets/figs/bt_no_cg.pdf}
%     \caption{ReAct saturates and fails to utilize additional tool budget, reaching a performance ceiling, while ReAct + Budget Tracker continues to scale effectively with larger budgets.}
%     \label{fig:bt_no_cg}
% \end{figure}

\textbf{Budget Tracker achieves lower resource usage and cost under similar accuracy.}
In Table~\ref{tab:react_less_cost}, the consumed resources report the average number of search and browse tool calls, as well as the unified cost combining tool usage and token consumption per question.
Increasing the tool-use budget in ReAct improves performance but also raises both tool call frequency and total cost. In contrast, Budget Tracker achieves comparable accuracy with less budget (10 vs.\ 100), while using 40.4\% fewer search calls, 19.9\% fewer browse calls, and reducing overall cost by 31.3\%. This demonstrates that explicit budget awareness enables more efficient tool use without sacrificing accuracy.

\begin{wrapfigure}{r}{0.48\linewidth}
    \centering
    \includegraphics[width=\linewidth]{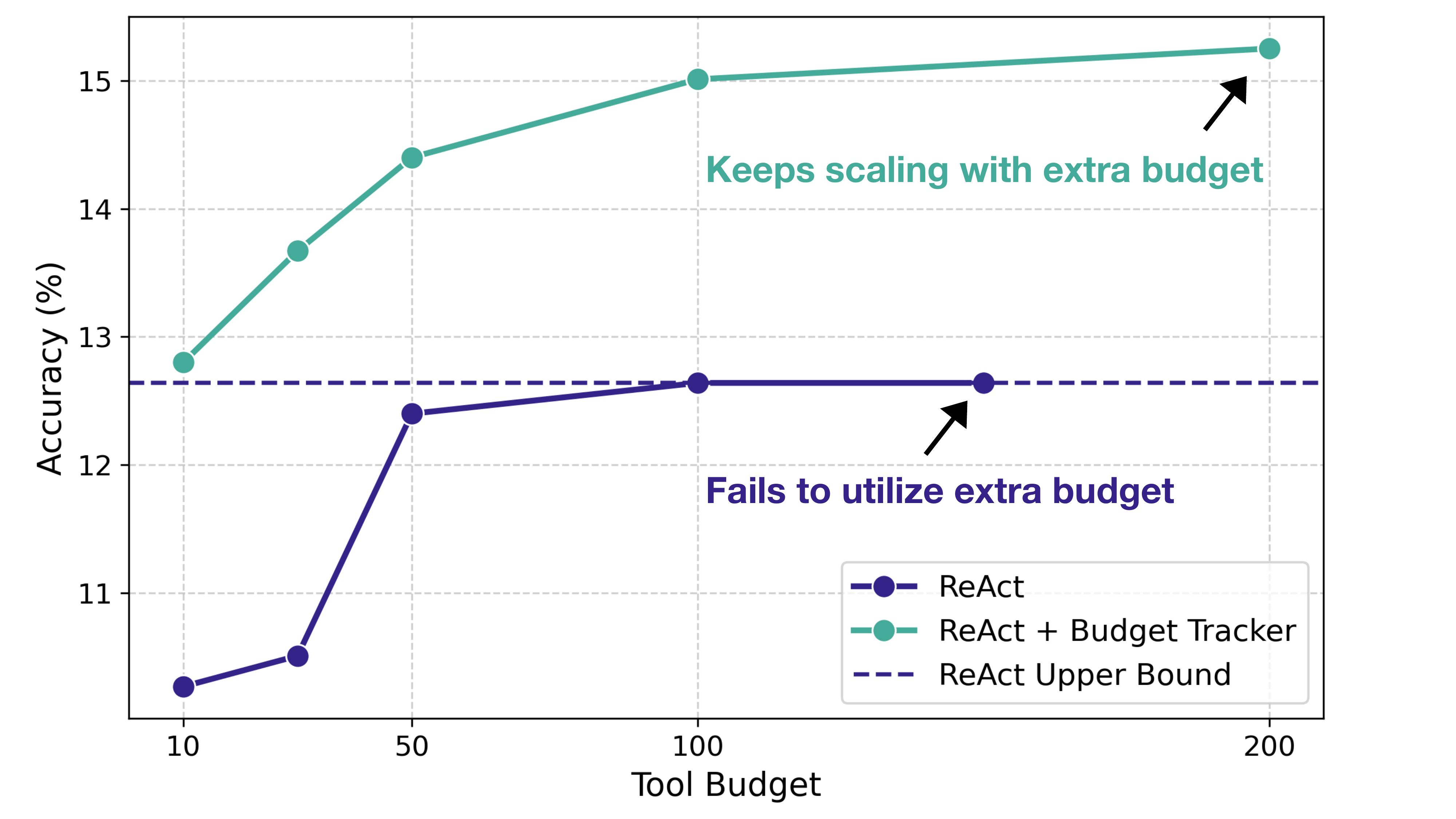}
    \caption{ReAct saturates and fails to utilize additional tool budget, reaching a performance ceiling, while Budget Tracker continues to scale effectively with larger budgets.}
    \label{fig:bt_no_cg}
\end{wrapfigure}

\textbf{Budget Tracker facilitates continued scaling by simply adding budget awareness.} Figure~\ref{fig:bt_no_cg} illustrates Gemini-2.5-Pro's performance on BrowseComp as tool budget increases. ReAct saturates at a budget of 100 on BrowseComp, terminating prematurely because it believes it has found a sufficient answer or is stuck, unaware of unused resources, even though the context window is not filled up. Budget Tracker breaks this ceiling by explicitly surfacing remaining resources, enabling the agent to leverage larger budgets and continue scaling.

% Lacking budget awareness, ReAct saturates at a budget of 100. Beyond this point, it fails to utilize any additional resource, even though the context window is not filled up.
% This ceiling occurs because the agent's reasoning process terminates prematurely: it either believes it has found a sufficient answer or concludes it is stuck and gives up, unaware of the unused resources.
% In contrast, Budget Tracker explicitly informs the model of its remaining resources, enabling it to leverage larger budgets and continue scaling its performance.

% \textbf{Budget Tracker facilitates continued scaling by simply adding budget awareness.} Figure~\ref{fig:bt_no_cg} shows that ReAct saturates at a budget of 100 on BrowseComp, terminating prematurely because it believes it has found a sufficient answer or is stuck, unaware of unused resources. Budget Tracker breaks this ceiling by explicitly surfacing remaining resources, enabling the agent to leverage larger budgets and continue scaling.

\subsection{Analysis on Test-time Scaling Strategies}

% \textbf{Budget Tracker consistently pushes the Pareto frontier and demonstrates robust effectiveness across different test-time scaling strategies.}
% Sequential scaling encourages the agent to use more tool calls by prompting it to continue exploring additional information sources. The prompting stops once the agent refuses to invoke any tools for five iterations. In Figure~\ref{fig:bt_seq_pro}, while this strategy enables ReAct to further improve accuracy, it eventually encounters a performance ceiling, where the agent confidently concludes that no further tool calls are necessary. In contrast, Budget Tracker achieves a consistently better scaling curve and sustains performance gains beyond ReAct's plateau. Furthermore, it advances the Pareto frontier of the cost--performance curve, demonstrating not only more effective performance scaling but also more efficient tool utilization.

\begin{figure}[t]
    \centering
    \begin{minipage}[t]{0.48\linewidth}
        \centering
        \includegraphics[width=\linewidth]{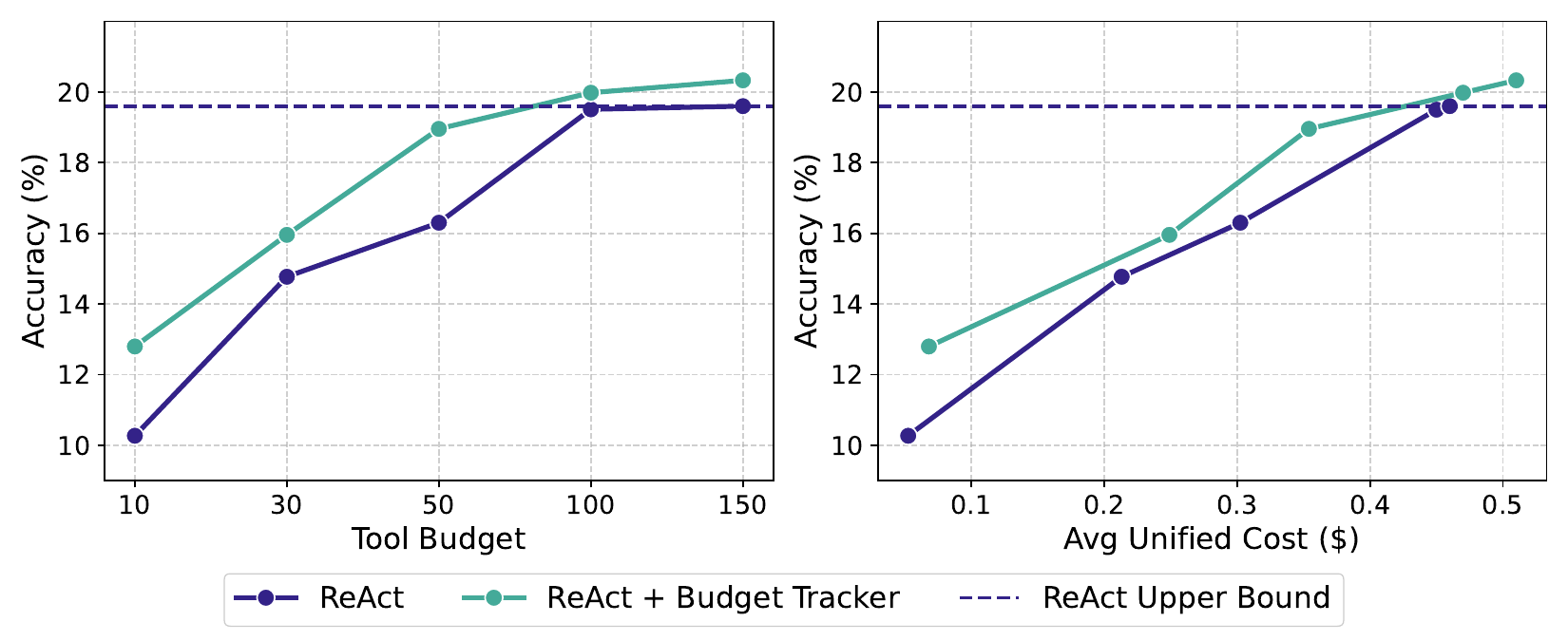}
        \caption{Sequential scaling comparison (Gemini-2.5-Pro). With explicit budget awareness, Budget Tracker overcomes early plateaus by adapting budget utilization to extend the cost-performance frontier.}
        \label{fig:bt_seq_pro}
    \end{minipage}
    \vspace{-5pt}
    \hfill
    \begin{minipage}[t]{0.48\linewidth}
        \centering
        \includegraphics[width=\linewidth]{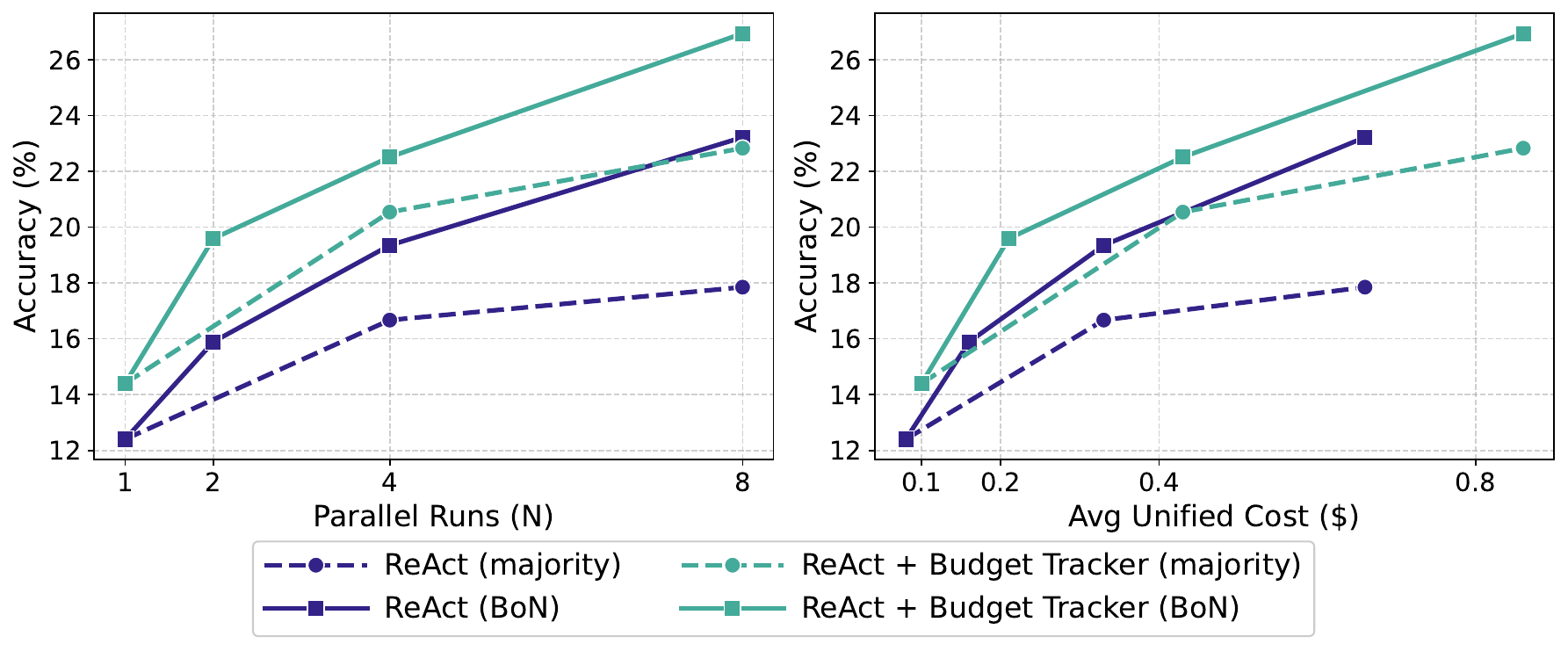}
        \caption{Parallel scaling comparison (Gemini-2.5-Pro). The left subfigure shows accuracy scaling with increasing parallel runs, while the right subfigure illustrates the corresponding cost–performance trend.}
        % \vspace{-5pt}
        \label{fig:bt_parallel_all}
    \end{minipage}

\end{figure}

\textbf{Budget Tracker consistently pushes the Pareto frontier across different test-time scaling strategies.}
Sequential scaling prompts the agent to continue exploring until it refuses to invoke any tools for five consecutive iterations, where the agent concludes that no tool calls are necessary. In Figure~\ref{fig:bt_seq_pro}, this strategy improves ReAct but eventually hits a ceiling. Budget Tracker sustains gains beyond this plateau and advances the cost--performance Pareto frontier, achieving both more effective scaling and more efficient tool utilization.

% \begin{figure}[h]
%     \centering
%     \includegraphics[width=0.5\linewidth]{assets/figs/bt_seq_pro.pdf}
%     \caption{Sequential scaling comparison (Gemini-2.5-Pro). With explicit budget awareness, Budget Tracker consistently outperforms ReAct, overcoming early plateaus by adapting budget utilization to extend the cost-performance frontier.
%     }
%     \label{fig:bt_seq_pro}
% \end{figure}

% Comparison of Budget Tracker and ReAct in sequential scaling using Gemini-2.5-Pro. With explicit budget awareness, Budget Tracker consistently improves upon ReAct at equal budgets. ReAct plateaus early as it cannot utilize extra resources, while Budget Tracker adapts its spending to gain further improvements and extend the cost-performance frontier.

For parallel scaling, we set a fixed budget of 50 per run for Gemini-2.5-Pro, as empirical observations indicate that only 0.8\% of the BrowseComp data requires more (see Appendix~\ref{app:usage_breakdown}). In Figure~\ref{fig:bt_parallel_all}, we use identical sample sizes for both settings and report Majority Vote and Best-of-N results (Pass@N results in Appendix~\ref{app:pass@n}). On the right, the figure illustrates the cost--performance scaling, where Budget Tracker consistently yields a superior curve.
% This proves that budget awareness is essential for pushing the Pareto frontier of test-time compute, regardless of the scaling strategy.

% \begin{figure}[t]
%     \centering
%     \includegraphics[width=0.5\linewidth]{assets/figs/bt_parallel_combined_bon.pdf}
%     \caption{Parallel scaling comparison (Gemini-2.5-Pro). The left subfigure shows accuracy scaling with increasing parallel runs, while the right subfigure illustrates the corresponding cost–performance trend.
%     }
%     \label{fig:bt_parallel_all}
%     % \vspace{-5pt}
% \end{figure}

\section{\method: Budget-Aware Test-time Scaling}

\begin{figure*}[h] 
    \centering
    \includegraphics[width=\textwidth]{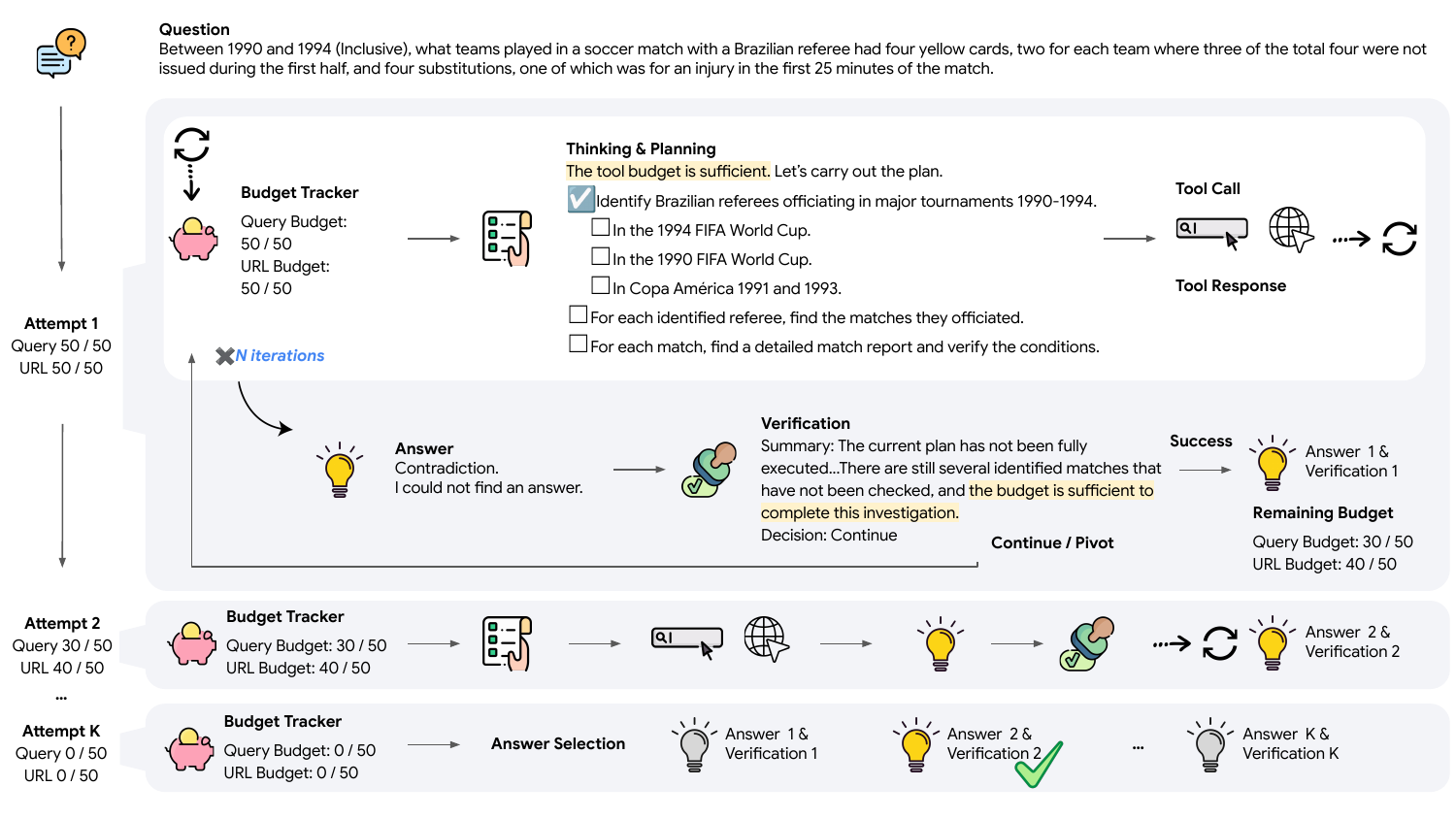} 
    \caption{Overview of \method. Given a question and per-tool budgets, the agent performs budget-aware thinking and planning via a checklist. It iterates by reasoning over new information and updated budgets. When an answer is proposed, verification decides to continue, pivot, or restart. \method terminates when any budget is exhausted.
    % Given a question and per-tool budgets, the agent begins with budget-aware thinking and planning, structured as a checklist. The agent keeps iterating by reasoning over new information and updated budgets. When an answer is proposed, \method performs verification and decides to either continue, pivot, or initiate a new attempt with the remaining budget. \method terminates when any of the budgets is exhausted.
    }
    \label{fig:framework}
\end{figure*}

% Given the effectiveness of budget awareness, we now explore how it enhances agent orchestration and influences key behaviors such as planning and self-verification.
% In this section, we propose \textbf{\method}, a \textbf{B}udget-\textbf{A}ware \textbf{T}est-time \textbf{S}caling framework for tool-augmented agents under explicit budget. 

Building on the effectiveness of budget awareness, we propose \textbf{\method}, a \textbf{B}udget-\textbf{A}ware \textbf{T}est-time \textbf{S}caling framework that we instantiate for search agents.
% We instantiate BATS for search agents, though its core principles are applicable to broader tool-use settings. 
As shown in Figure~\ref{fig:framework}, given a question and a tool call budget, \method begins with internal reasoning to formulate a structured action plan and decide which tools to invoke.
Tool calls are triggered through special tokens~\citep{DBLP:conf/nips/HaoLWH23}, and the returned tool responses are appended to the reasoning sequence, expanding the context with new evidence.  
When the agent proposes a candidate answer, the verification module checks its validity and decides whether to continue the current sequence or initiate a new attempt with the remaining budget. The iterative process terminates once any budgeted resource is exhausted. Finally, an LLM-as-a-judge selects the best answer across all verified candidates. 
The complete algorithm pseudocode is in Appendix~\ref{app:algorithm}; the full system instructions are provided in Appendix~\ref{app:prompts}.

\subsection{Budget-Aware Planning}

Planning in \method incorporates both \emph{constraint decomposition} and \emph{structured dynamic planning}. 
Selecting a good starting point is critical: a well-chosen entry narrows the search space and conserves budget, while a poor choice quickly exhausts it (see example in Appendix~\ref{app:case_plan}).
To address this, the agent first decomposes question clues into two categories: (1) \emph{exploration}, to expand the candidate space, and (2) \emph{verification}, to validate specific properties. While a verification clue can sometimes provide a direct shortcut to the answer, relying on it prematurely is risky, as it may consume resources without guaranteeing progress. 

The agent maintains a persistent, \textbf{tree-structured plan} that acts as a dynamic checklist for resource allocation, where all completed, failed, or partial steps are retained to prevent redundant tool calls (full prompt in Appendix~\ref{app:prompt_plan}).
In Figure~\ref{fig:framework}, each step represents a subtask requiring multiple tool calls. After each iteration, the module refines the tree by adding branches, resolving steps, or invalidating unproductive paths, adjusting exploration breadth and verification depth based on the remaining budget.

\subsection{Budget-Aware Self-verification}
% answer verification, decision, summary
Once the agent proposes an answer, the self-verification module re-evaluates the reasoning trajectory and corresponding resource usage (full prompt in Appendix~\ref{app:prompt_verification}). 
It performs a constraint-by-constraint backward check: using the extracted exploration and verification clues, the module assesses each constraint as satisfied, contradicted, or unverifiable, grounding the answer directly against the question's requirements.

Based on this analysis, the module then makes a verification decision considering budget status. If all constraints are satisfied, the answer is marked as a \textit{SUCCESS}. If several constraints remain unverifiable but the trajectory is promising and budget permits deeper exploration, the outcome is \textit{CONTINUE}. 
If contradictions are found or remaining budget cannot support further investigation, the agent terminates low-yield directions early and \textit{PIVOT}s to avoid wasting resources while alternatives remain viable.

% In contrast, if contradictions are identified or the remaining budget cannot support further investigation towards this lead, the agent is expected to terminate expensive or low-yield directions early and \textit{PIVOT} toward a different direction to avoid wasting tool call resources while resources are still sufficient to pursue alternatives.

When the decision is to \textit{CONTINUE} or \textit{PIVOT}, the module generates a concise summary replacing the raw trajectory in context, including key reasoning steps, intermediate findings, failure causes, and suggestions for optimization to avoid redundant exploration.
By compressing the reasoning trajectory into a compact and informative summary, the verifier reduces context length while keeping subsequent steps grounded in prior findings.

% Together, structured constraint checking, explicit decision rules, and budget-aware summarization allow \method to terminate useless trajectories early, continue promising ones efficiently, and maintain progress toward the correct answer within strict budget constraints.

Together, constraint checking, explicit decision rules, and budget-aware summarization allow \method to terminate low-yield trajectories early and continue promising ones efficiently.

% Together, structured constraint checking, explicit decision rules, and budget-aware trajectory summarization allow \method to terminate useless trajectories early, continue promising ones efficiently, and maintain reliable progress toward the correct answer within strict budget constraints. 

% Experimental setup is included in results.tex in the COLM 2026 version.
% \input{google_chapters_26/experiments}

\section{Experiments and Results}

\subsection{Setup}\label{sec:exp}

\paragraph{Datasets.} We evaluate our methods on 3 information seeking benchmarks: BrowseComp~\citep{DBLP:journals/corr/abs-2504-12516}, a dataset of 1,266 web-browsing questions requiring persistent retrieval; BrowseComp-ZH~\citep{DBLP:journals/corr/abs-2504-19314}, a 289-question Chinese benchmark designed to test agents’ performance in region-specific web environments; and HLE-Search~\citep{DBLP:journals/corr/abs-2507-16075}, a curated 200-question subset of Humanity's Last Exam~\citep{DBLP:journals/corr/abs-2501-14249} focusing on queries that explicitly require search rather than pure reasoning. We also evaluate Budget Tracker on 114 tasks from $\tau^2$-bench~\citep{DBLP:journals/corr/abs-2506-07982} retail subset which focuses on domain-specific user-agent conversations. The details are provided in Appendix~\ref{app:tau2}.

% We compare \method against a range of models and agentic frameworks, including both general-purpose base models~\citep{DBLP:journals/corr/abs-2410-21276, DBLP:journals/corr/abs-2412-16720, gemini25, anthropic2025claude37sonnet, anthropic2025claude4} and those specifically fine-tuned for agentic search tasks~\citep{DBLP:journals/corr/abs-2507-02592, Liu2025WebExplorerEA, DBLP:journals/corr/abs-2508-07976,lu2025deepdiveadvancingdeepsearch}. Specifically, \textbf{SLIM} periodically summarizes the running trajectory into a concise state that is carried forward to facilitate longer horizons and more efficient scaling~\citep{DBLP:journals/corr/abs-2510-18939}. To evaluate the final answer accuracy, we use Gemini-2.5-Flash as the judge model and adopt the evaluation prompt from~\citet{DBLP:journals/corr/abs-2501-14249}. Baseline results for HLE-Search are scraped from~\citet{DBLP:journals/corr/abs-2507-16075}.

\paragraph{Baselines.}
We compare \method against a range of models and agentic frameworks, including both general-purpose base models called directly without tool augmentation~\citep{DBLP:journals/corr/abs-2412-16720, gemini25, anthropic2025claude37sonnet, anthropic2025claude4} and those specifically fine-tuned for agentic search tasks~\citep{DBLP:journals/corr/abs-2507-02592, Liu2025WebExplorerEA}. Specifically, \textbf{SLIM} periodically summarizes the running trajectory into a concise state that is carried forward to facilitate longer horizons and more efficient scaling~\citep{DBLP:journals/corr/abs-2510-18939}.
To evaluate the final answer accuracy, we use Gemini-2.5-Flash as the judge model and adopt the evaluation prompt from~\citet{DBLP:journals/corr/abs-2501-14249}. Baseline results for HLE-Search are scraped from~\citet{DBLP:journals/corr/abs-2507-16075}. We use a temperature of 0.7 for execution and 0.0 for evaluation; implementation details and temperature analysis are in Appendix~\ref{sec:implementation} and~\ref{app:temperature}.

% To evaluate the final answer accuracy, we use Gemini-2.5-Flash as the judge model and adopt the evaluation prompt from~\citet{DBLP:journals/corr/abs-2501-14249}. Baseline results for HLE-Search are scraped from~\citet{DBLP:journals/corr/abs-2507-16075}.

To analyze scaling behavior, we primarily compare \method with parallel scaling. This ensures a direct, resource-equivalent comparison: increasing parallel sample size scales total tool consumption in the same way \method’s iterative attempts do. By default, we use Gemini-2.5-Flash to select the most common answer via a majority vote~\citep{DBLP:conf/iclr/0002WSLCNCZ23} (see Appendix~\ref{app:pass@n} for details). For experiments in Section~\ref{sec:main_results}, to enforce the budget constraint for each question, we continue sampling new sequences until the budget is fully consumed. Thus the number of sampled sequences may differ across questions.

% To effectively scale the number of tool calls and analyze agent's behaviors under varying tool call budget, we mainly compare \method with parallel scaling as increasing the sample size directly scales the total tool consumption. This is also similar to how \method tries a new attempt. By default, we use Gemini-2.5-Flash to select the most common answer via a majority vote~\citep{DBLP:conf/iclr/0002WSLCNCZ23} (see Appendix~\ref{app:pass@n} for details). For experiments in \method, to enforce the budget constraint for each question, we continue sampling new sequences until the budget is fully consumed. Thus the number of sampled sequences may differ across questions.

% For scaling methods, we evaluate two approaches applied to the ReAct~\citep{DBLP:conf/iclr/YaoZYDSN023}.
% For \textbf{sequential scaling}, we append the instruction~\citep{DBLP:journals/corr/abs-2501-19393} that encourages it to rethink and use more tools whenever the agent provides an answer. This process is repeated until the tool budget is exhausted, after which it is prompted to produce the final answer.
% For \textbf{parallel scaling}, we use temperature sampling to generate diverse responses. By default, we use Gemini-2.5-Flash to select the most common answer via a majority vote~\citep{DBLP:conf/iclr/0002WSLCNCZ23} (see Appendix~\ref{app:pass@n} for details). For experiments in \method, to enforce the budget constraint for each question, we continue sampling new sequences until the budget is fully consumed. Thus the number of sampled sequences may differ across questions.

\subsection{Results}\label{sec:main_results}

\begin{wrapfigure}{r}{0.48\linewidth}
    \centering
    \includegraphics[width=\linewidth]{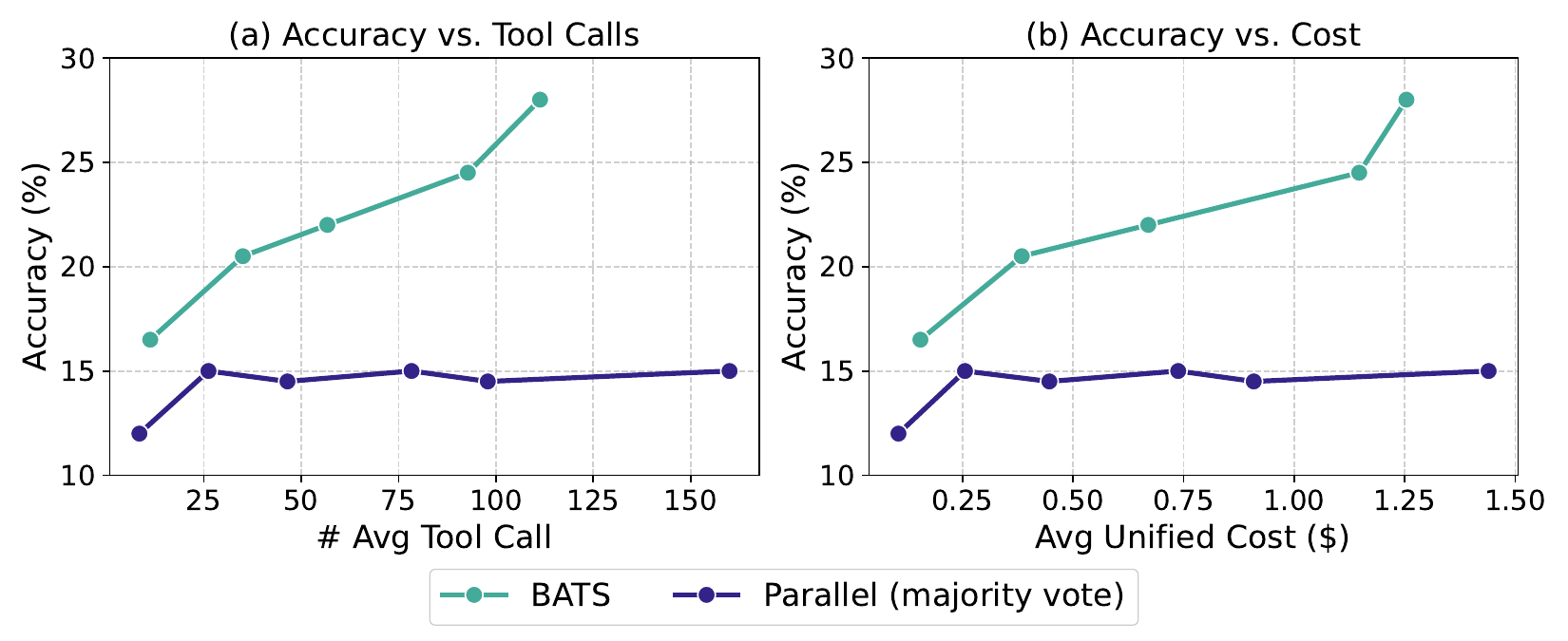}
    \caption{Scaling behaviors along (a) total number of tool calls and (b) average unified cost, evaluated on a 200-example subset of BrowseComp using Gemini-2.5-Pro.}
    \label{fig:two}
    % \vspace{-10pt}
\end{wrapfigure}

Table~\ref{tab:main} shows the performance comparison across different web search agents. 
Under the strict budget constraints of 100 tool uses for BrowseComp, \method consistently achieves better results than baselines, obtaining 24.6\% on BrowseComp, 46.0\% on BrowseComp-ZH and 27.0\% on HLE-Search using Gemini-2.5-Pro. Crucially, while many competing agents rely on extensive task-specific training to boost performance, \method is entirely training-free. This result highlights the effectiveness of our budget-aware design, which maximizes the efficiency of every tool call to achieve superior results in resource-constrained settings without the need for additional fine-tuning.

\begin{table*}[t!]
\centering
\scalebox{0.9}{
\begin{tabular}{lrrrr}
\toprule
\textbf{Method} & \textbf{Training} & \textbf{BrowseComp} & \textbf{BrowseComp-ZH} & \textbf{HLE-Search}\\ 
\midrule
\multicolumn{5}{l}{\textit{Model Only (w/o tools)}} \\ \midrule
% GPT-4o & \xmark & 0.6 & 6.2  & - \\
Claude-3.7-Sonnet$^\star$ & \xmark & 2.3 & 11.8  & - \\
Gemini-2.5-Flash & \xmark & 2.7 & 23.9  & 2.8 \\
Gemini-2.5-Pro & \xmark & 6.3 & 27.8  & 8.6 \\
% OpenAI o1 & \xmark & 9.9 & 29.1  & - \\
\midrule
\multicolumn{5}{l}{\textit{Training-based Agents (w/ tools)}} \\ \midrule
% ASearcher & \cmark & 5.2 & 15.6  & - \\
WebSailor$^\star$ & \cmark & 12.0 & 30.1  & - \\
% DeepDive & \cmark & 14.8 & 25.6  & - \\
WebExplorer$^\star$ & \cmark & 15.7 & 32.0  & - \\
OpenAI Deep Research$^\star$ & \cmark & \textbf{51.5} & \textbf{42.9}  & \textbf{29.1} \\
\midrule
\multicolumn{5}{l}{\textit{Budget-constrained (w/ tools)}} \\ \midrule

Claude-Sonnet-4 \hspace*{12pt} ReAct & \xmark & 12.2 & 29.1 & 20.5 \\
Claude-Sonnet-4 \hspace*{12pt} SLIM & \xmark & 11.0 & 25.3 & 15.0 \\

\rowcolor{gray!20}
Claude-Sonnet-4  \hspace*{12pt} \method (Ours) & \xmark & 19.1 & 41.5 & \textbf{29.0} \\

Gemini-2.5-Flash \hspace*{10pt} ReAct & \xmark & 9.7 & 26.5 & 14.7 \\
Gemini-2.5-Flash \hspace*{10pt} SLIM & \xmark & 10.6 & 26.3 & 14.5 \\
%todo: flash results!
\rowcolor{gray!20}
Gemini-2.5-Flash  \hspace*{10pt} \method (Ours) & \xmark & 14.3 & 34.3 & 19.5 \\

Gemini-2.5-Pro \hspace*{18pt} ReAct & \xmark & 12.6 & 31.5 & 20.5 \\
Gemini-2.5-Pro \hspace*{18pt} SLIM & \xmark & 13.4 & 31.5 & 14.5 \\

\rowcolor{gray!20}
Gemini-2.5-Pro \hspace*{18pt} \method (Ours) & \xmark & \textbf{24.6} & \textbf{46.0} & 27.0 \\
% ---------------------------------------------------------

\bottomrule
\end{tabular}
}
\caption{Performance comparison across web search agents. Results cited from other publications are marked with $^\star$. The ``Training'' column specifies whether the model has been specifically trained on agentic web search tasks. For our budget-constrained setting, each agent is provided a budget of 100 tool uses per tool.}
\label{tab:main}
\end{table*}

\textbf{\method achieves higher performance under the same budget constraint.} We vary the tool-call budget and evaluate performance on a random subset of 200 examples from BrowseComp for a manageable analysis.
Figure~\ref{fig:two} (left) shows the accuracy against the average number of tool calls, including both search and browse.
Across all budget levels, \method consistently outperforms the parallel majority-vote baseline, demonstrating that budget-aware adaptation leads to more effective use of limited tool calls.

\textbf{\method achieves higher performance when accounting for unified costs.}
Beyond tool-call counts, we measure performance under a unified cost metric that incorporates both token usage and tool-call expenses. In Figure~\ref{fig:two} (right), \method achieves more favorable scaling curves, delivering higher accuracy at comparable or lower costs. \method not only improves effectiveness under budget constraints but also yields better cost–performance trade-offs.

% \textbf{Additional analysis.} We provide module ablations (Appendix~\ref{app:ablation}), resource consumption breakdowns (Appendix~\ref{app:bats_resource}), and case studies (Appendix~\ref{app:case_plan}) in the Appendix.

\subsection{Analysis}\label{app:ablation}

\begin{table}[h]
\centering

\scalebox{0.9}{
\begin{tabular}{lrrr}
\toprule
\textbf{Method} & \textbf{BrowseComp} & \textbf{BrowseComp-ZH} & \textbf{HLE-Search} \\
\midrule
BATS             & \textbf{18.7} & \textbf{39.1} & \textbf{23.0} \\
\quad w/o Planning     & 17.0 & 34.6 & 20.0 \\
\quad w/o Verification & 15.4 & 37.7 & 22.0 \\
\quad w/o Planning \& Verification  & 14.6  & 32.9   & 21.5 \\
\bottomrule
\end{tabular}
}
\caption{Ablation results of BATS with Gemini-2.5-Pro (budget = 100 per tool). Removing different modules leads to various performance drops across datasets.}
\label{tab:ablations}
% \vspace{-5pt}
\end{table}

\textbf{Ablations.} We present an ablation analysis of the key modules within \method using Gemini-2.5-Pro in Table~\ref{tab:ablations}. Across all settings, the agent is allocated a budget of 100 tool uses per tool. To provide a clearer view of the orchestration framework, we focus on \method with the early-stopping mechanism, where generation terminates once an answer is verified as success.
The gap between these results and Table~\ref{tab:main} demonstrates that continuing to initiate new attempts with the remaining budget after an initial verified answer leads to substantially higher accuracy.

Results indicate that removing the planning leads to a moderate performance decrease. However, removing the verification module causes a more significant drop on BrowseComp (from 18.7\% to 15.4\%). This suggests that the verification module is crucial for enabling the agent to navigate the solution space and accurately assess its current progress. Removing both modules results in lower performance (14.6\% on BrowseComp), while BrowseComp-ZH and HLE-Search exhibit some variance, likely due to the smaller dataset size. In addition, questions in HLE-Search are typically shorter and contain fewer details that can be verified, which limits the contribution of the verification module. Overall, the trend confirms that both modules contribute positively to the orchestration framework's effectiveness.

\textbf{Additional analysis.} In the Appendix, we further analyze a single-attempt early stopping setting where the agent may terminate before exhausting its budget (\ref{app:early}), resource consumption breakdowns (\ref{app:bats_resource}), and case studies (\ref{app:case_study}). We also provide an analysis of the behavioral mechanisms underlying these improvements in Appendix~\ref{app:why}.

\section{Related Work}

\subsection{Test-time Scaling}
Test-time scaling (TTS)~\citep{DBLP:journals/corr/abs-2408-03314, DBLP:journals/corr/abs-2503-24235} strategies include sequential refinement~\citep{DBLP:conf/nips/MadaanTGHGW0DPY23, DBLP:journals/tmlr/ZhangYYY25,DBLP:journals/corr/abs-2501-19393, DBLP:journals/corr/abs-2507-14295}, parallel sampling with aggregation~\citep{DBLP:journals/corr/abs-2407-21787,DBLP:conf/iclr/0002WSLCNCZ23}, and hybrid approaches~\citep{DBLP:journals/corr/abs-2504-00762,DBLP:journals/corr/abs-2501-19306,DBLP:conf/naacl/WanWCL25,DBLP:conf/iclr/LiYFPW0W024}. While these focus on text-only reasoning, we extend TTS to tool-augmented agents, where scaling accounts for both tokens and tool calls under budget constraints.
A complementary line of work constrains the scaling effort via tokens, sampled sequences, or FLOPs~\citep{DBLP:journals/corr/abs-2407-19825,DBLP:journals/tmlr/WelleckBFSXNKH24,DBLP:conf/iclr/DamaniSPBA25,DBLP:journals/corr/abs-2504-13367}. AgentTTS~\citep{DBLP:journals/corr/abs-2508-00890} optimizes LLM size and sampling under a unified FLOPs budget, while SLIM~\citep{DBLP:journals/corr/abs-2510-18939} uses periodic summarization to manage context growth.
In contrast, we formalize and constrain the tool-call budget, shifting the focus to budget-constrained scaling of tool-augmented agents.

\subsection{Web Search Agents}

Web search agents use search and browse tools to solve complex, multi-hop queries~\citep{DBLP:journals/corr/abs-2508-06600,DBLP:journals/corr/abs-2508-07999,team2025kimi,DBLP:journals/corr/abs-2507-16075,DBLP:journals/corr/abs-2510-24701}. 
One research direction builds training data and applies various training methods to specialize the models~\citep{DBLP:journals/corr/abs-2503-09516,DBLP:journals/corr/abs-2505-22648,DBLP:journals/corr/abs-2507-02592,Liu2025WebExplorerEA,DBLP:journals/corr/abs-2507-15061,DBLP:journals/corr/abs-2510-24699,DBLP:journals/corr/abs-2508-13167}.
MiroThinker~\citep{miromindteam2025mirothinkerpushingperformanceboundaries} shows scaling interactive tool use is an effective dimension, but it often yields redundant or inefficient tool calls.
Another explores inference-time strategies~\citep{DBLP:journals/corr/abs-2501-05366,DBLP:journals/corr/abs-2506-15741,Qiao2025WebResearcherUU,DBLP:journals/corr/abs-2509-25301}, such as incorporating programmatic execution to perform multiple tool call actions~\citep{DBLP:journals/corr/abs-2508-09129}, finding an optimal, statically efficient configuration~\citep{DBLP:journals/corr/abs-2508-02694}, or exploring various design choices when scaling test-time compute~\citep{DBLP:journals/corr/abs-2506-12928}. Instead, our work focuses on dynamic, cost-effective performance, providing the first analysis of agent scaling behavior under explicit budget constraints.

\section{Conclusion}
In this work, we present the first systematic study of budget-constrained tool-use agents and their test-time scaling behaviors.
We identify that standard agents often hit a performance ceiling due to a lack of resource awareness. To overcome this, we introduce the \textbf{Budget Tracker}, a lightweight plug-in that provides budget awareness, and \textbf{\method}, a comprehensive framework that dynamically adapts planning and verification based on real-time resource status. Extensive experiments across multiple models and information-seeking benchmarks show that budget awareness enables stronger agent scaling and consistently pushes the cost-performance Pareto frontier. By formalizing tool-call budgets as a critical scaling dimension, our work offers a principled foundation for building efficient and adaptive agents.

\section*{Acknowledgement}

We thank Rui Meng, and members from Google Cloud AI Research for their valuable
feedback during the preparation of the paper.

\bibliography{main}

\appendix

\section{Experiment Details}

\subsection{Resource Cost}
The cost of tool calls is determined by the pricing of the providers. 
To standardize billing, we've established a unified rate of \$0.001 per invocation for both search API calls and web browsing actions. 
Because different URLs produce varying amounts of text and tokens, this per-call rate is derived as an average computed from post hoc statistics over all our experiments.
The consumption of tokens is billed separately, adhering to the official pricing models of the API provider pricing\footnote{\url{https://cloud.google.com/vertex-ai/generative-ai/pricing}}.

\subsection{Implementation Details}\label{sec:implementation}

% \subsection{Implementation Details}
We use Gemini-2.5-Flash, Gemini-2.5-Pro~\citep{gemini25} and Claude-Sonnet-4~\citep{anthropic2025claude4} as the backbone models in our framework. By default, we disable thinking for Gemini-2.5-Flash by setting the thinking budget to 0. For Gemini-2.5-Pro, we set the thinking budget to 1024. The maximum number of new tokens for generation was set to 65,536 for Gemini models and 64,000 for Claude. We use a temperature of 0.7 during agent execution to encourage exploration, and use a deterministic temperature of 0.0 for final answer selection and answer evaluation. We use the Google Custom Search JSON API\footnote{\url{https://developers.google.com/custom-search/v1/overview}} and SERPER\footnote{\url{https://serper.dev/}}for search tools, Jina.ai\footnote{\url{https://jina.ai/}} and Crawl4AI~\citep{crawl4ai2024} for web browsing.
When reproducing SLIM~\citep{DBLP:journals/corr/abs-2510-18939}, we set the maximum number of iterations to 100, the maximum token count to 64K, and the summary interval to 50 following the default settings. 

To keep the context length manageable, we employ several simple context management strategies. For each browse tool call, the fetched webpage is truncated to 150,000 characters before being sent to the model. At every iteration, we discard tool outputs from previous steps and retain only the most recent tool response, preventing the context from growing with accumulated tool results. 
In \method's verification module, we further control context size by periodically replacing the historical trajectory with summary. Since the agent determines when to activate the verification module, we perform a check during each invocation: if more than $K$ iterations have passed since the last update (with $K=10$ in our experiments), we replace the older reasoning trace with a concise summary derived from the verification outputs.

\section{Algorithm}\label{app:algorithm}

Algorithm~\ref{alg:bats} provides the complete pseudocode for \method. Given a question $x$ and per-tool budgets $\mathbf{b}=(b_1,\dots,b_K)$, the agent iteratively reasons, plans, and invokes tools while the Budget Tracker maintains real-time resource states. When a candidate answer is proposed, the self-verification module evaluates it and decides to \textit{CONTINUE}, \textit{PIVOT}, or declare \textit{SUCCESS}. Upon pivoting, the trajectory is compressed into a summary and a new attempt begins with the remaining budget. The process terminates when any budget is exhausted, and a final judge selects the best answer among all verified candidates.

\begin{algorithm}[th]
\caption{\method: Budget-Aware Test-time Scaling}
\label{alg:bats}
\begin{algorithmic}[1]
\Require Question $x$, tool set $\mathcal{T}=\{t_1,\dots,t_K\}$, per-tool budgets $\mathbf{b}=(b_1,\dots,b_K)$
\Ensure Final answer $\hat{y}$
\State $\mathcal{A} \gets \emptyset$ \Comment{Verified answer set}
\State $\mathbf{u} \gets (0,\dots,0)$ \Comment{Per-tool usage counters}
\State $\mathcal{S} \gets \emptyset$ \Comment{Trajectory summaries from previous attempts}
\State $\hat{y}_{\text{last}} \gets \texttt{None}$ \Comment{Most recent candidate answer}
\State Decompose $x$ into exploration and verification constraints
\State Initialize tree-structured plan $\mathcal{P}$
\While{$u_i < b_i$ for all $i = 1,\dots,K$} \Comment{Budget not exhausted}
    \State \textbf{// Budget-Aware Planning \& Reasoning}
    \State Inject budget status $(\mathbf{b} - \mathbf{u})$ via Budget Tracker
    \State Generate reasoning conditioned on $x$, $\mathcal{P}$, $\mathcal{S}$, and budget status
    \State Update plan $\mathcal{P}$: add branches, resolve or invalidate steps
    \If{agent decides to invoke tool $t_i$ with arguments $a$}
        \State $r \gets t_i(a)$ \Comment{Execute tool call}
        \State $u_i \gets u_i + 1$
        \State Append response $r$ to reasoning context
    \ElsIf{agent proposes candidate answer $\hat{y}_c$}
        \State $\hat{y}_{\text{last}} \gets \hat{y}_c$
        \State \textbf{// Budget-Aware Self-Verification}
        \State $d \gets \textsc{Verify}(\hat{y}_c, x, \text{trajectory}, \mathbf{b} - \mathbf{u})$
        \If{$d = \textsc{Success}$}
            \State $\mathcal{A} \gets \mathcal{A} \cup \{\hat{y}_c\}$
            \State Summarize trajectory $\to s$; \ $\mathcal{S} \gets \mathcal{S} \cup \{s\}$
            \State Reset reasoning context; start new attempt
        \ElsIf{$d = \textsc{Continue}$}
            \State Compress historical trajectory if context threshold is exceeded
            \State Continue current reasoning path
        \ElsIf{$d = \textsc{Pivot}$}
            \State Summarize trajectory $\to s$; \ $\mathcal{S} \gets \mathcal{S} \cup \{s\}$
            \State Reset reasoning context; abandon current direction in $\mathcal{P}$
        \EndIf
    \EndIf
\EndWhile
\If{$\mathcal{A} \neq \emptyset$}
    \State $\hat{y} \gets \textsc{JudgeSelect}(\mathcal{A})$ \Comment{LLM-as-judge picks best answer}
\Else
    \State $\hat{y} \gets \hat{y}_{\text{last}}$ \Comment{Fallback to last candidate}
\EndIf
\State \Return $\hat{y}$
\end{algorithmic}
\end{algorithm}

\section{Additional Analysis}

\subsection{Tool Usage Breakdown}\label{app:usage_breakdown}

\begin{table*}[h]
\centering
\small
\begin{tabular}{lrrrrrr}
\toprule
\multirow{2}{*}{\textbf{Model}} &
\multirow{2}{*}{\textbf{Budget}} &
\multirow{2}{*}{\textbf{Acc \%}} &
\multicolumn{2}{c}{\textbf{Query}} &
\multicolumn{2}{c}{\textbf{URL}} \\
\cmidrule(lr){4-5} \cmidrule(lr){6-7}
& & & \textbf{Avg. \#} & \textbf{Over-Budget \%} 
  & \textbf{Avg. \#} & \textbf{Over-Budget \%} \\
\midrule

% ====================== Pro ======================
\multirow{3}{*}{Gemini-2.5-Pro}
& 30  & 10.5 & 13.77 & 9.24   & 1.33 & 0.00 \\
& 50  & 12.4 & 13.95 & 0.79   & 1.31 & 0.00 \\
& 100 & 12.6 & 14.24 & 0.00   & 1.36 & 0.00 \\

\midrule

% ====================== Flash ======================
\multirow{3}{*}{Gemini-2.5-Flash}
& 30  & 9.0  & 22.77  & 42.81  & 1.04 & 0.00 \\
& 50  & 9.7 & 28.88 & 14.22  & 1.24 & 0.00 \\
& 100 & 10.0 & 29.93 & 2.60   & 1.36 & 0.00 \\

\bottomrule
\end{tabular}
\caption{Resource usage statistics of the ReAct baseline using Gemini-2.5-Pro and Gemini-2.5-Flash on the BrowseComp dataset. Over-Budget \% denotes the percentage of data that reaches the given tool budget limit.}
\label{tab:react_budget_usage}
\end{table*}

In Table~\ref{tab:react_budget_usage}, we examine how the ReAct baseline utilizes tools under different budget constraints on the BrowseComp dataset. As expected, increasing the allowed tool budget enables the agent to interact more with the environment, leading to improved accuracy. The Over-Budget\% column shows the proportion of examples that exhaust the allocated tool budget. For Gemini-2.5-Pro, only 0.8\% of the data requires more than 50 tool calls, whereas Gemini-2.5-Flash shows a higher reliance on tool usage, with 2.6\% of the queries exceeding a budget of 100. Notably, Flash models issue roughly twice as many search queries as Pro models. This suggests that the stronger parametric knowledge of Gemini-2.5-Pro allows it to navigate the search space more efficiently, reducing reliance on externally gathered evidence and enabling it to solve queries with fewer tool calls.

\subsection{Generalization to Conversational Agents}\label{app:tau2}

To verify the generalizability of budget awareness beyond search agents, we extend our experiments to general conversational tasks via $\tau^2$-bench~\citep{DBLP:journals/corr/abs-2506-07982}. Unlike search agents, agents in this benchmark must engage in multi-turn dialogues with users and invoke various tools to resolve complex, real-world tasks.

\textbf{Experiment Details.} A key advantage of our Budget Tracker plugin is its high modularity and ease of integration into existing open-source frameworks. We implement all experiments in this section by plugging the tracker into the official $\tau^2$-bench public codebase\footnote{\url{https://github.com/sierra-research/tau2-bench}} with minimal modifications.
For the experimental setup, we utilize Gemini-2.5-Flash (32K context window) as the backbone for both the agent and the user simulator, with temperatures set to 0.7 and 0.0, respectively. To reflect realistic resource constraints, we apply a unified global budget rather than per-tool limits. We set total tool-call limits of 10, 30, and 50, allowing the agent to flexibly allocate its available resources across each task execution. We use the Retail subset, which contains 114 tasks in total.

\textbf{Metrics.} We use the metrics from the official codebase and report Avg@$k$ and Pass\textasciicircum $k$ scores~\citep{DBLP:journals/corr/abs-2406-12045, DBLP:journals/corr/abs-2506-07982}. Pass\textasciicircum $k$ reflects the rate at which a task succeeds in all $k$ independent attempts, averaged across the task set. This metric highlights reliability and consistency, which are important for conversational customer service tasks in this dataset.

\textbf{Results.} Table~\ref{tab:tau2_results} demonstrates the effectiveness of budget awareness across two key dimensions. First, \textbf{under the same budget constraints, Budget Tracker consistently delivers better performance.} At a budget of 10, it improves Avg@4 from 47.8\% to 55.5\% and Pass\textasciicircum 4 from 21.9\% to 27.2\%. Second, \textbf{under similar or higher performance levels, Budget Tracker utilizes lower costs and fewer tool calls}. Specifically, our method at a budget of 10 already surpasses the Baseline's Pass\textasciicircum 4 performance at a budget of 50 (27.2\% vs.\ 26.3\%), while requiring approximately 18.4\% fewer tool calls (5.38 vs.\ 6.59) and lower total cost. This indicates that budget-aware agents achieve higher reliability through strategic resource management rather than simply relying on increased tool call invocations.

\begin{table}[h]
\centering

% \resizebox{\linewidth}{!}{
\begin{tabular}{lrrrrr}
\toprule
\multirow{2}{*}{\textbf{Method}} & 
\multirow{2}{*}{\textbf{Budget}} & 
\multicolumn{2}{c}{\textbf{Performance (\%)}} & 
\multicolumn{2}{c}{\textbf{Consumed resources}} \\ 
\cmidrule(lr){3-4} \cmidrule(lr){5-6}
 &  & \textbf{Avg@4} & \textbf{Pass\textasciicircum 4} & \textbf{\# tool} & \textbf{cost (¢)} \\
\midrule
\multirow{3}{*}{Baseline} 
 & 10 & 47.8 & 21.9 & 6.29 & 2.64 \\
 & 30 & 59.0 & 26.3 & 6.59 & 2.89 \\
 & 50 & 59.0 & 26.3 & 6.59 & 2.89 \\
\midrule
\multirow{3}{*}{+ Budget Tracker} 
 & 10 & 55.5 & 27.2 & 5.38 & 2.72 \\
 & 30 & 61.4 & 28.9 & 6.66 & 3.00 \\
 & 50 & \textbf{63.2} & \textbf{36.0} & 6.82 & 3.04 \\
\bottomrule
\end{tabular}
% }
\caption{Generalizability evaluation on conversational tasks via $\tau^2$-bench Retail subset using Gemini-2.5-Flash. Under unified global budgets (10, 30, 50 tool-call limits), the modular Budget Tracker consistently enhances reliability (Pass\textasciicircum 4) and average performance (Avg@4) compared to the baseline.}
\label{tab:tau2_results}
\end{table}

\subsection{Generalization to Coding Agents}\label{app:swebench}

To further evaluate whether budget awareness generalizes beyond search and conversational settings, we conduct an additional experiment on all 500 tasks in SWE-bench Verified~\citep{jimenez2024swebench} using Gemini-3-Flash-Preview within the mini-swe-agent framework~\citep{yang2024sweagent}. Coding agents in this setting must inspect and modify software repositories through iterative Bash tool use, providing a substantially different tool-use environment from the web search tasks studied in the main paper.

\textbf{Experiment Details.} We integrate Budget Tracker into mini-swe-agent and use the same budget-awareness prompt as in our main experiments, with only minor wording changes that adapt the instructions from search and browse tools to Bash tool use. We impose a global tool-call budget for each task and compare the standard baseline against the baseline augmented with Budget Tracker.

\textbf{Metrics.} We report the resolve rate on SWE-bench Verified, the mean number of tool calls consumed per task, and the mean token cost in cents. Table~\ref{tab:swebench_results} summarizes the results.

\begin{table}[h]
\centering
\begin{tabular}{lrrrr}
\toprule
\multirow{2}{*}{\textbf{Method}} &
\multirow{2}{*}{\textbf{Budget}} &
\multicolumn{1}{c}{\textbf{Performance (\%)}} &
\multicolumn{2}{c}{\textbf{Consumed resources}} \\
\cmidrule(lr){3-3} \cmidrule(lr){4-5}
& & \textbf{Resolve Rate} & \textbf{\# tool} & \textbf{cost (¢)} \\
\midrule
\multirow{3}{*}{Baseline}
& 50 & 36.8 & 46.0 & 18.50 \\
& 60 & 53.2 & 50.7 & 20.40 \\
& 70 & 63.0 & 53.6 & 21.59 \\
\midrule
\multirow{2}{*}{+ Budget Tracker}
& 50 & 55.4 & 45.7 & 18.88 \\
& 70 & \textbf{70.0} & 55.1 & 23.31 \\
\bottomrule
\end{tabular}
\caption{Generalizability evaluation on all 500 SWE-bench Verified tasks using Gemini-3-Flash-Preview within the mini-swe-agent framework. Budget Tracker improves resolve rate while maintaining comparable tool usage and token cost.}
\label{tab:swebench_results}
\end{table}

\textbf{Results.} Table~\ref{tab:swebench_results} shows that the benefits of budget awareness extend to software engineering agents. First, \textbf{under the same budget constraints, Budget Tracker substantially improves task resolution at comparable resource usage.} At a budget of 50, Budget Tracker increases the resolve rate by 18.6 percentage points (55.4\% vs.\ 36.8\%) while using a similar number of tool calls (45.7 vs.\ 46.0) and incurring a similar token cost (18.88¢ vs.\ 18.50¢).

Second, \textbf{Budget Tracker produces a clear Pareto improvement over a higher-budget baseline.} Budget Tracker with a budget of 50 outperforms the baseline with a budget of 60 on every measured axis: it uses 9.9\% fewer tool calls (45.7 vs.\ 50.7), reduces token cost by 7.5\% (18.88¢ vs.\ 20.40¢), and improves resolve rate by 2.2 percentage points (55.4\% vs.\ 53.2\%). These results reinforce our main conclusion: across web search, conversational, and coding tasks, explicit budget awareness helps agents make more effective use of available tool calls across diverse tool-use domains.

\subsection{Parallel Scaling Details}\label{app:pass@n}

For Majority Vote, we aggregate the final answers across different samples and use a judge model (Gemini-2.5-Flash) to identify the consensus answer. For Best-of-N, we provide all response trajectories to the judge model and select the most promising one. For Pass@N, the score is calculated as 1 if any of the sampled responses contain the correct answer.
The prompts used for these evaluations are provided in Appendix~\ref{app:prompt_answer}.
In Figure~\ref{fig:bt_parallel_pass}, we additionally report the Pass@N results for ReAct and Budget Tracker under parallel scaling settings. Budget Tracker consistently achieves higher overall accuracy across varying budgets and cost levels.

\begin{figure}[h]
    \centering
    \includegraphics[width=0.8\linewidth]{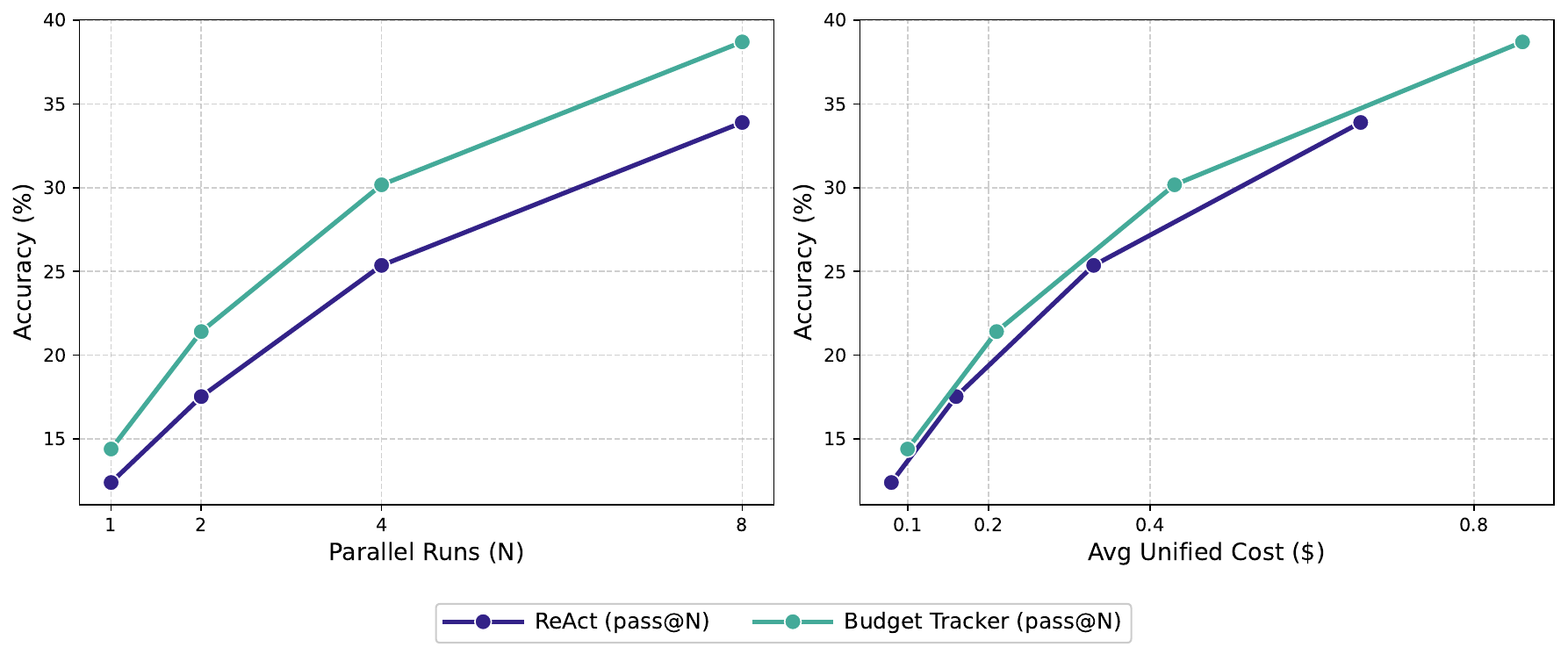}
    \caption{Pass@N results of Budget Tracker and ReAct in parallel scaling. 
The left subfigure shows accuracy scaling with increasing parallel runs, 
while the right subfigure illustrates the corresponding cost–performance trend.
    }
    \label{fig:bt_parallel_pass}
\end{figure}

\subsection{Temperature Sensitivity}\label{app:temperature}

For model generation, we adopt a temperature of $T = 0.7$ for all tasks, which facilitates the diverse reasoning paths necessary for exploration. For final answer selection, we use a temperature of $T = 0.0$ to maintain deterministic scoring and ensure reproducibility.
To validate these hyperparameter choices, we conduct a comparison using a 200-sample subset of the BrowseComp dataset with Gemini-2.5-Flash. As shown in Table~\ref{tab:temperature_ablation}, the model exhibits stability across various decoding configurations, with $T = 0.7$ providing the optimal balance of exploration and precision.

\begin{table}[h]
    \centering
    
    \begin{tabular}{lcccc}
        \toprule
        \textbf{Temperature} & 0.0 & 0.3 & \textbf{0.7} & 1.0 \\
        \midrule
        \textbf{Acc (\%)} & 15.0 & 14.0 & \textbf{16.0} & 14.0 \\
        \bottomrule
    \end{tabular}
    \caption{Comparison of different temperature settings on model accuracy. Results are based on a 200-sample subset of BrowseComp using Gemini-2.5-Flash.}
    \label{tab:temperature_ablation}
\end{table}

% The marginal gains observed at $T = 0.7$ suggest that while the model is robust to temperature fluctuations, slight stochasticity aids in navigating complex browsing trajectories compared to purely greedy decoding.

% We follow the common practice of using temperature = 0.7 for agent tasks. For evaluation, we use temperature = 0.0 for answer selection to ensure deterministic scoring. To assess its impact, we conduct an additional experiment varying the temperature on a 200-sample subset of BrowseComp using Gemini-2.5-Flash. As shown below, the performance is relatively stable across settings, and temperature = 0.7 yields the best accuracy:

\subsection{Early Stopping Analysis}\label{app:early}

% \subsection{Early Stopping}

We analyze how effectively agents can solve questions under budget constraints when allowed to stop early, without exhausting all tool calls. This setting reflects realistic scenarios where efficient, confident answers are preferable to prolonged exploration.
To isolate this behavior, we focus on the agent’s \textit{first attempt} only. For \method, this is the first answer that successfully passes self-verification. For baselines that lack verification, we use the output from a single generation pass as their first attempt.
If any tool budget is exhausted before this first attempt completes, the agent will stop and output a final answer based on its progress so far. This evaluation captures both efficiency and robustness, testing whether agents can solve questions quickly without being forced to use their full budgets.

\begin{figure}[h]
    \centering
    \includegraphics[width=0.6\linewidth]{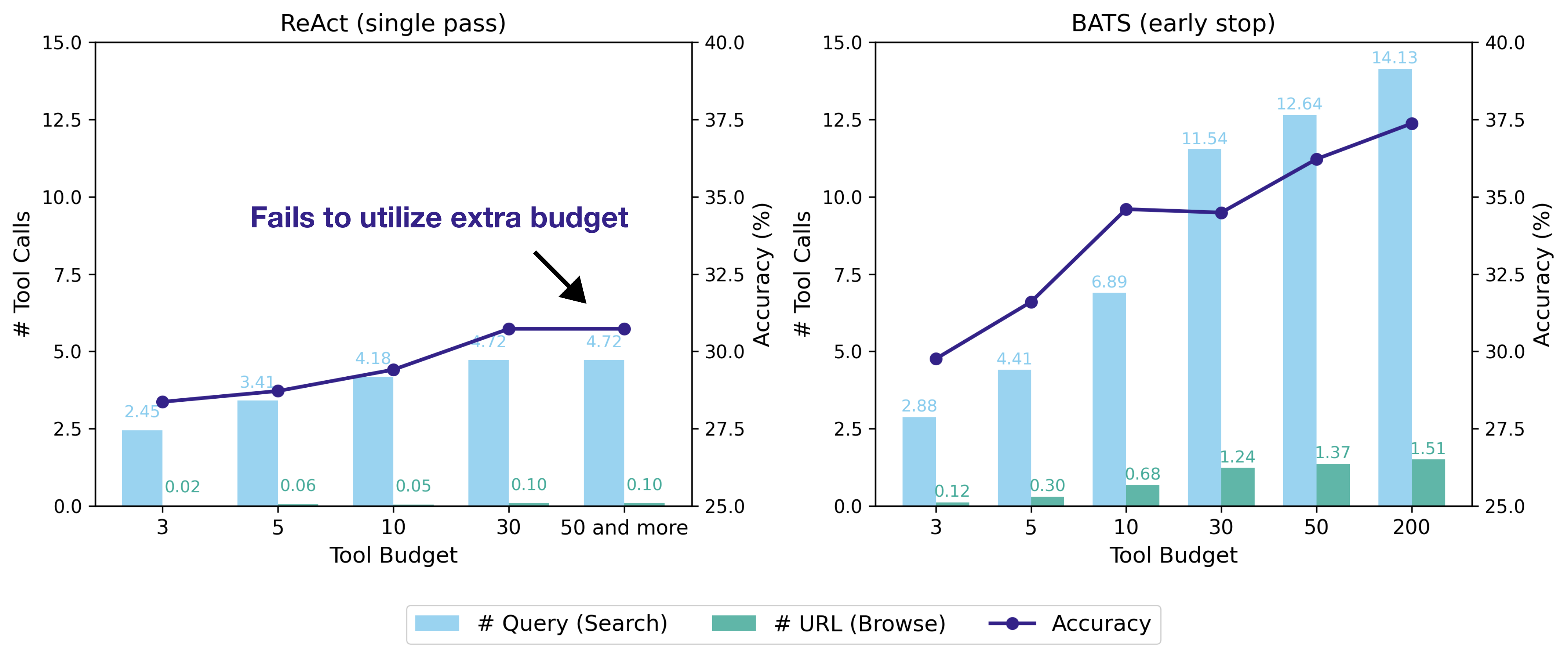}
    \caption{Performance and resource utilization with early stopping on BrowseComp-ZH using Gemini-2.5-Pro. The bars represent the average number of Search and Browse tool calls (left axis), and the line plots indicate accuracy (right axis). While ReAct plateaus and fails to utilize additional budget, \method effectively scales up tool usage to achieve higher accuracy as the budget increases.}
    \label{fig:first-bczh}
    % \vspace{-5pt}
\end{figure}

% \begin{wrapfigure}{r}{0.5\linewidth}
%     \centering
%     \includegraphics[width=\linewidth]{assets/figs/early_stop_bczh_pro.pdf}
%     \caption{Performance and resource utilization with early stopping on BrowseComp-ZH using Gemini-2.5-Pro. The bars represent the average number of Search and Browse tool calls (left axis), and the line plots indicate accuracy (right axis). While ReAct plateaus and fails to utilize additional budget, \method effectively scales up tool usage to achieve higher accuracy as the budget increases.}
%     \label{fig:first-bczh}
%     \vspace{-5pt}
% \end{wrapfigure}

\textbf{Budget awareness enables \method to scale effectively with increased resources.} Figure~\ref{fig:first-bczh} presents the early stopping performance on BrowseComp-ZH using Gemini-2.5-Pro. The x-axis denotes the predefined tool call budget for both search and browse tool calls. Bars show the average number of tool calls actually used, while the line plot reports the resulting accuracy.
The ReAct baseline (Figure~\ref{fig:first-bczh} left) exhibits poor budget utilization: it consistently underuses the browse tool (fewer than 0.1 calls on average) and reaches a performance plateau early, with accuracy capped at 30.7\% for all budgets of 30 and above. This behavior shows its inability to benefit from additional resources.
In contrast, \method (Figure~\ref{fig:first-bczh} right) shows strong budget-adaptive behavior. As the budget increases, it strategically raises its use of both search and browse tools, leading to steady gains from 29.8\% (budget=3) to 37.4\% (budget=200). 
Notably, with a budget of 5, \method already surpasses the baseline’s best achievable accuracy. This demonstrates \method's ability to make more strategic and cost-effective decisions, delivering higher accuracy even with fewer resources. 
% More analysis can be found in Appendix~\ref{app:early}.

\begin{wrapfigure}{r}{0.5\textwidth}  
  \centering
  \includegraphics[width=0.95\linewidth]{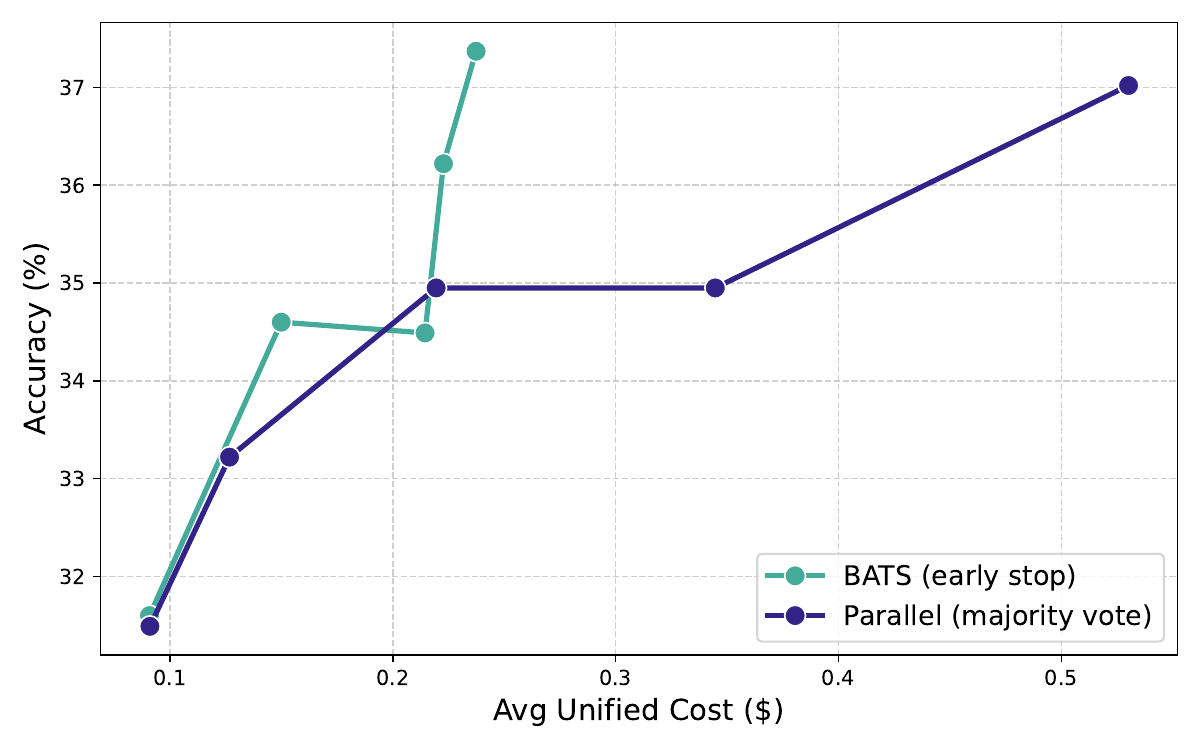}
  \caption{Average unified cost analysis on BrowseComp-ZH using Gemini-2.5-Pro.}
  \label{fig:first-bczh-cost}
\end{wrapfigure}

\textbf{\method pushes the cost–performance Pareto frontier even further by leveraging early stopping.}
To provide a direct and transparent comparison of cost-efficiency, Figure~\ref{fig:first-bczh-cost} shows accuracy against the actual unified cost. \method demonstrates a much steeper performance curve, indicating that it achieves higher accuracy for a lower cost compared to the parallel majority vote baseline. \method reaches over 37\% accuracy for approximately \$0.23, while the parallel baseline requires more than double that cost (over \$0.50) to achieve a comparable result.
This efficiency benefits from its budget-aware verification module, which enables early termination upon finding a satisfactory answer and minimizes unnecessary spending.

\subsection{Resource Usage Breakdown}\label{app:bats_resource}

Table~\ref{tab:resource_detail} reports the detailed breakdown of tool calls and token consumption in \method using Gemini-2.5-Pro on BrowseComp. We compare our approach against the variant \method-response, which keeps the full tool responses from previous rounds. In contrast, \method removes these responses to optimize context length.
In the table, token usage is reported as a triplet: (Input / Output / Cache). The results demonstrate that removing historical tool responses does not compromise performance (maintaining similar accuracy) but significantly reduces computational overhead. This efficiency is evidenced by the substantial decrease in cache tokens and the reduction in overall cost.

\begin{table}[h]
    \centering
    
    \small
    \begin{tabular}{lcccccc}
        \toprule
        & \textbf{Acc (\%)} & \textbf{\# search} &\textbf{\# browse} & \textbf{Total tokens} & \textbf{Veri. tokens} & \textbf{Cost (\$)} \\
        \midrule
        BATS & 24.6 & 87.3 & 13.6 & 32.1 / 6.9 / 39.3 & 1.1 / 1.2 / 7.6 & 1.1 \\
        BATS-response & 24.3 & 84.4 & 14.6 & 33.6 / 6.5 / 91.8 & 1.0 / 1.3 / 17.1 & 1.2 \\
        \bottomrule
    \end{tabular}
    \caption{Resource usage comparison. Token counts are reported as Input / Output / Cache (in 10,000). \method achieves comparable accuracy to the full-context baseline (\method-response) while significantly reducing cache token usage and cost.}
    \label{tab:resource_detail}
\end{table}

% \subsection{Planning Module}
% To evaluate the impact of the planning module, we augment the ReAct baseline with our proposed design, which integrates constraint analysis and a dynamically structured checklist plan. The tool-call budget is capped at 200 for both search queries and browse URLs. As shown in Table~\ref{tab:abl_plan}, the addition of planning module alone improves the agent’s ability to organize exploration and utilize tool usage more effectively, resulting in a performance gain of 1.8\%. 

% \begin{table}[h]
% \centering
% \caption{Effect of the planning module. With the same budget, the planning module encourages better tool utilization and yields higher average performance.}
% \begin{tabular}{lrrr}
% \toprule
% \textbf{Method} & \textbf{Acc \%} & \textbf{Avg. \# Query} & \textbf{Avg. \# URL} \\
% \midrule
% ReAct & 11.0 & 7.75 & 0.35 \\
% ReAct + planning & 12.8 & 13.81 & 0.82 \\
% \bottomrule
% \end{tabular}
% \label{tab:abl_plan}
% \end{table}

\lstdefinestyle{boxedcode}{
  basicstyle=\ttfamily\scriptsize,
  columns=fullflexible,
  keepspaces=true,
  breaklines=true,
  upquote=true,
  lineskip=-1pt
}

\section{Prompts}\label{app:prompts}

Our prompt for web search agents is developed based on \citet{DBLP:journals/corr/abs-2503-09516} and \citet{DBLP:journals/corr/abs-2507-02592}.

\subsection{ReAct+ Budget Tracker}\label{app:prompt_budget}

\begin{center} 
\begin{tcolorbox}[
  enhanced,
  breakable,                 
  colback=white,
  colframe=black!50!white,
  title={ReAct + Budget Tracker},
  width=1.0\linewidth        
]
\lstset{style=boxedcode}
\begin{lstlisting}
You are an AI reasoner with Google Search and Browsing tools. Solve the question by iterating: think, tool_code, tool_response, answer.

## Tools
You have access to 2 tools: search and browse.
{
  "name": "search",
  "description": "Performs batched web searches: supply an array 'query'; the tool retrieves the top 10 results for each query in one call.",
  "parameters": {
    "type": "object",
    "properties": {
      "query": {
        "type": "array",
        "items": {
          "type": "string"
        },
        "description": "Array of query strings. Include multiple complementary search queries in a single call."
      }
    },
    "required": [
      "query"
    ]
    }
},
{
  "name": "browse",
    "description": "Visit webpage(s) and return the summary of the content.",
    "parameters": {
        "type": "object",
        "properties": {
            "url": {
                "type": "array",
                "items": {"type": "string"},
                "description": "The URL(s) of the webpage(s) to visit. Can be a single URL or an array of URLs."
            },
            "goal": {
                "type": "string",
                "description": "The specific information goal for browsing webpage(s)."
            }
        },
        "required": [
            "url",
            "goal"
        ]
    }
}

You should start with one or more cycles of (thinking about which tool to use -> performing tool code -> waiting for tool response), and end with (thinking about the answer -> answer of the question). The thinking processes, tool codes, tool responses, and answer are enclosed within their tags. There could be multiple thinking processes, tool codes, tool call parameters and tool response parameters.

## Budget

You have two independent budgets:
- Query Budget (for search)
- URL Budget (for browse)
Each string in 'query' or 'url' consumes 1 unit respectively.

After each <tool_response>, a <budget> tag shows remaining units.
You must ADAPT your strategy dynamically to the current budget state.

### HIGH Budget (>=70% remaining)
- Search: 3-5 diverse queries in one batch.
- Browse: up to 2-3 high-value URLs.
- Goal: Broad exploration, build context fast.

### MEDIUM Budget (30%-70%)
- Search: 2-3 precise, refined queries per cycle.
- Browse: 1-2 URLs that close key knowledge gaps.
- Goal: Converge; eliminate uncertainty efficiently.

### LOW Budget (10%-30%)
- Search: 1 tightly focused query.
- Browse: at most 1 most promising URL.
- Goal: Verify a single critical fact or finalize answer.

### CRITICAL (<10% remaining or 0 in one budget)
- Avoid using the depleted tool.
- Only perform 1 minimal-cost query or browse if absolutely essential.
- If uncertainty remains and no tool use is possible, output <answer>None</answer>.

## Step syntax

<think>
Thinking process. Analyze the query, your internal knowledge, and search results to build your reasoning. Always justify tool choices based on the remaining budgets.
</think>
<tool_code>
{"name": "tool name here", "arguments": {"parameter name here": parameter value here, "another parameter name here": another parameter value here, ...}}
</tool_code>
<tool_response>
tool_response here
</tool_response>
<budget>
Query Budget Used: [number], Query Budget Remaining: [number], URL Budget Used: [number], URL Budget Remaining: [number]
</budget>

Repeat <think><tool_code> until you have the final answer.
<answer> Final solution only. </answer> 

## About answers

* Only write the final answer inside <answer> and </answer>.
* If you cannot find the answer, write <answer>None</answer>.
\end{lstlisting}
\end{tcolorbox}
\captionsetup{justification=centering} 
\captionof{figure}{Prompts used in ReAct + Budget Tracker.}
\label{fig:prompt_budegt}
\end{center}

\subsection{Planning Module}\label{app:prompt_plan}

\begin{center} % 仅用于整体居中（可选）
\begin{tcolorbox}[
  enhanced,
  breakable,                 % 允许跨页
  colback=white,
  colframe=black!50!white,
  title={Budget-aware Planning Module in BATS},
  width=1.0\linewidth        % 控制盒子宽度
]
\lstset{style=boxedcode}
\begin{lstlisting}
## About questions

Questions contain two types of constraints: exploration and verification. 
* Exploration: Broad, core requirements (e.g., birthday, profession). Use these for initial searches to surface candidates. You may combine 1-2 to form stronger queries.
* Verification: Narrow, specific details. Apply these only after you have candidates, to confirm or filter them. Never begin with verification constraints.
Start with exploration queries, then use verification to validate the results.

## About planning

Maintain a tree-structured checklist of actionable steps (each may require several tool calls).
- Mark each step with its status: [ ] pending, [x] done, [!] failed, [~] partial.
- Use numbered branches (1.1, 1.2) to represent alternative paths or candidate leads.
- Log resource usage after execution: (Query=#, URL=#).
- Keep all executed steps, never delete them, retain history to avoid repeats.
- Update dynamically as you reason and gather info, adding or revising steps as needed.
- Always consider current and remaining budget when updating the plan.
\end{lstlisting}
\end{tcolorbox}
\captionsetup{justification=centering} % 可选：让标题居中
\captionof{figure}{Prompts used in planning module of BATS.}
\label{fig:prompt_plan}
\end{center}

\subsection{Self-verification Module}\label{app:prompt_verification}

\begin{center} % 仅用于整体居中（可选）
\begin{tcolorbox}[
  enhanced,
  breakable,                 % 允许跨页
  colback=white,
  colframe=black!50!white,
  title={Budget-aware Self-verification Module in BATS},
  width=1.0\linewidth        % 控制盒子宽度
]
\lstset{style=boxedcode}
\begin{lstlisting}
You are an AI Strategic Verifier. Your primary goal is to evaluate a proposed answer, assess the viability of the current problem-solving plan, and decide the best course of action: declare success, continue with the current plan, or pivot to a new one.

### Given Inputs

* Question: The original user question. An answer is believed to exist.
* Trajectory: The sequence of reasoning steps and tool calls taken so far in the current attempt.
* Current Answer: The final answer produced by the current attempt.
* Budget Status: Information on current tool call budget utilization and remaining budget, including search queries and browsing urls.

### Your Task: A 3-Step Process

You must proceed in the following order:

#### Step 1: Conduct Verification Analysis

First, perform a strict verification of the `Current Answer`.
* Go through each constraint from the original Question one by one.
* For each constraint, compare it against the `Current Answer` and the `Trajectory`.
* State your finding for each constraint: satisfied, contradicted, or unverifiable.

#### Step 2: Make a Strategic Decision

Based on your verification and the budget, make one of three decisions.

1. SUCCESS: If the verification in Step 1 passed (all constraints are satisfied). The task is complete.
2. CONTINUE: If the verification failed because few constraints are unverifiable, but the overall plan is still sound and salvageable. This is the choice if **both** of these conditions are true:
    * Promising Path: The `Trajectory` is generally sound, and the failure was due to a correctable error.
    * Sufficient Budget: There is enough `Remaining Budget` to attempt a correction on this path.
3. PIVOT: If the verification failed, signal to abandon the current plan and switch to another one. You should pivot if any of these conditions are true:
    * Dead End: The `Trajectory` reveals a fundamental flaw in the current plan's logic that cannot be easily fixed.
    * Failed Tool Calls: The Trajectory shows repeated, unsuccessful attempts to find certain info.
    * Insufficient Budget: The `Remaining Budget` is too low to make another meaningful attempt or correction within the *current* plan.

#### Step 3: Summarize for the Next Step

This is the most critical step for guiding future actions. 
You need to first provide a **trajectory summary**: Summarize the agent's reasoning trajectory into a concise narrative. Explain its initial goal, the logical steps taken, key findings and the final conclusion, emphasizing how key findings or contradictions caused the agent to change its strategy.

Then, provide additional details tailored to your decision in Step 2.

* If the decision is SUCCESS:
    * No further detail needed.

* If the decision is CONTINUE / PIVOT:
    * Failure Analysis: Diagnose the root cause of the failure. Identify the critical flaw (e.g., poor query design, flawed logic, misinterpreted evidence) and name the general failure pattern to prevent its recurrence.
    * Useful information: Any useful intermediate findings or results from the current `Trajectory` that could be valuable inputs for the next attempt. This prevents redundant work.
    * Strategic Recommendations: Provide actionable advice for the agent's next attempt. Suggest strategic pivots, new angles of investigation, or different ways to combine the problem's constraints. Explicitly state if it should backtrack to and resume from a specific step in the previous plan to avoid re-doing work.

### **Output Requirement**

Your final output must be a single JSON object with the following structure. Do not add any text before or after this JSON block.

```json
{
  "verification": "Verification analysis",
  "decision": "SUCCESS | CONTINUE | PIVOT",
  "justification": "A concise explanation for your strategic decision. Why is it a success, a dead end, or a correctable error?",
  "trajectory_summary": "The informative trajectory summary.",
  "details": "A JSON object containing the additional details required by Step 3. For a SUCCESS decision, this can be an empty object {}."
}
```
\end{lstlisting}
\end{tcolorbox}
\captionsetup{justification=centering} % 可选：让标题居中
\captionof{figure}{Prompts used in self-verification module of BATS.}
\label{fig:prompt_verification}
\end{center}

\subsection{Answer Selection}\label{app:prompt_answer}

\begin{center} % 仅用于整体居中（可选）
\begin{tcolorbox}[
  enhanced,
  breakable,                 % 允许跨页
  colback=white,
  colframe=black!50!white,
  title={Best-of-N Answer Selection},
  width=1.0\linewidth        % 控制盒子宽度
]
\lstset{style=boxedcode}
\begin{lstlisting}
You are an expert evaluator. Your task is to select the most accurate and specific answer to an information-seeking question. The question has a deterministic answer. You'll be provided with several answers and their corresponding trajectories/verifications.

**Instructions:**
1.  **Identify the Core Question:** Determine the exact piece of information the question is asking for (e.g., a person, a location, a date).
2.  **Evaluate Candidates:** For each candidate phrase, assess its factual accuracy.
3.  **Compare and Select:** Choose the answer that is more likely to be correct. You should never choose "None" as the answer.

**Output Format:**
First, provide a brief justification explaining why the chosen answer is the most accurate and specific choice. Then, on a new line, output the letter of the best option inside a box.

**Example:**
Justification: Answer B is the most specific correct location...
Answer: \\boxed{B}
\end{lstlisting}
\end{tcolorbox}

\begin{tcolorbox}[
  enhanced,
  breakable,                 % 允许跨页
  colback=white,
  colframe=black!50!white,
  title={Majority Vote Answer Selection},
  width=1.0\linewidth        % 控制盒子宽度
]
\lstset{style=boxedcode}
\begin{lstlisting}
You are an expert evaluator. Your task is to select the answer that best represents the **majority vote** among the provided candidates. The question has a deterministic answer, and the goal is to identify which option most responses converge on. You'll be provided with several answers and their corresponding trajectories/verifications.

**Instructions:**
1. **Identify the Core Question:** Determine the exact piece of information the question is asking for (e.g., a person, a location, a date).  
2. **Tally the Votes:** Review all candidate answers and count how many times each distinct answer (or near-equivalent variant) appears. Treat semantically equivalent responses as votes for the same candidate.  
3. **Select the Majority:** Choose the answer that has the highest number of votes. If there is a tie, pick the option that is the most specific and consistent with the question. Never choose "None" or refuse to make a choice.  

**Output Format:**  
First, provide a brief justification explaining why the chosen answer was selected (e.g., ``Answer C has the majority of votes across candidates''). Then, on a new line, output the letter of the best option inside a box.  

**Example:**  
Justification: Answer B received the majority of votes and aligns most consistently with the question.  
Answer: \\boxed{B}
\end{lstlisting}
\end{tcolorbox}

\captionsetup{justification=centering} % 可选：让标题居中
\captionof{figure}{Prompts used in answer selection.}
\label{fig:prompt_ablation}
\end{center}

\section{Why Does Budget Awareness Work?}\label{app:why}

The previous sections demonstrate that budget awareness improves performance; here we consolidate evidence to explain \emph{why}. The key insight is not that budget-aware agents use more tools, but that they manage resources \emph{more strategically}---adapting their search strategy to the available budget and making higher-quality decisions about \emph{what} to do with each tool call.

\subsection{The Default Failure Mode: Rigid Strategy Regardless of Budget}

Standard agents apply a fixed behavioral pattern irrespective of the resource budget. Table~\ref{tab:react_budget_usage} shows that Gemini-2.5-Pro with ReAct uses nearly identical numbers of tool calls whether the budget is 30 or 100 (13.77 vs.\ 14.24 search calls), and its accuracy barely changes. The agent follows the same rigid strategy, issuing a fixed set of queries and terminating upon finding a plausible answer, without any awareness of how many resources remain or whether deeper investigation is warranted. This rigidity, rather than the raw quantity of tool calls, is the fundamental bottleneck.

\subsection{Behavioral Shift 1: Adaptive Strategy Selection}

Budget awareness enables agents to condition their search strategy on the current resource state, choosing qualitatively different approaches under different budget regimes.

The case studies in Figures~\ref{fig:case_high_bgt} and~\ref{fig:case_low_bgt} provide direct evidence. Under high budget, the budget-aware agent explicitly reasons about its resource abundance and selects a broad, exploratory strategy. Under low budget, the same agent shifts to a precision-oriented approach, crafting highly targeted queries and prioritizing the most promising leads. In contrast, the ReAct baseline uses the same brute-force, narrow-first strategy regardless of available resources, which happens to be suboptimal in both regimes: too narrow when resources are plentiful (missing relevant evidence), and too wasteful when resources are scarce (exhausting budget on low-yield queries).

This adaptivity is the primary driver behind Table~\ref{tab:react_less_cost}'s result: Budget Tracker with budget=10 achieves comparable accuracy to ReAct with budget=100 (12.8\% vs.\ 12.6\%) while using 40.4\% fewer search calls and reducing cost by 31.3\%. The improvement comes not from spending more, but from spending differently.

\subsection{Behavioral Shift 2: More Balanced Tool Allocation}

Beyond adapting the overall strategy, budget-aware agents allocate resources more effectively across tool types. From Table~\ref{tab:react_less_cost}, we compute the browse-to-search ratio as a proxy for exploration depth:
\begin{itemize}
    \item ReAct (Acc$=12.6\%$): $1.36 / 14.24 = 9.6\%$
    \item Budget Tracker (Acc=12.8\%): $1.09 / 8.48 = 12.9\%$
\end{itemize}

Budget Tracker devotes a proportionally larger share of its tool calls to browsing (deep reading of full webpages) rather than surface-level search. This is a meaningful shift: browse calls retrieve detailed content that search snippets cannot provide and are critical for answering the verification-heavy questions in BrowseComp. The budget-aware agent, knowing its remaining resources, makes more deliberate decisions about when a query result warrants deeper investigation, rather than defaulting to repeated shallow searches.

\subsection{Behavioral Shift 3: Budget-Conditioned Verification in \method}

\method's verification module further leverages budget awareness for strategic decision making. Rather than blindly continuing or stopping, it chooses among \textit{SUCCESS}, \textit{CONTINUE}, and \textit{PIVOT} based on both answer quality and remaining resources.

% [TODO] From your BATS trajectory logs (Gemini-2.5-Pro, budget=100, BrowseComp),
% compute and fill in the following table:

% \begin{table}[h]
% \centering
% \caption{Verification decision distribution in \method (Gemini-2.5-Pro, budget=100, BrowseComp).}
% \label{tab:verification_dist}
% \begin{tabular}{lr}
% \toprule
% \textbf{Metric} & \textbf{Value} \\
% \midrule
% Avg. \# verification triggers per question & [TODO] \\
% Avg. \# attempts per question & [TODO] \\
% \textsc{Success} rate & [TODO]\% \\
% \textsc{Continue} rate & [TODO]\% \\
% \textsc{Pivot} rate & [TODO]\% \\
% \bottomrule
% \end{tabular}
% \end{table}

This budget-conditioned decision making is critical for resource efficiency. The gap between \method's early-stopping accuracy (18.7\%, Table~\ref{tab:ablations}) and its full accuracy (24.6\%, Table~\ref{tab:main}) shows that verification-guided retries contribute a 5.9 percentage-point gain. Importantly, these retries are not blind repetitions: as illustrated in Figure~\ref{fig:case_verification}, the verifier diagnoses the specific failure cause and, recognizing sufficient remaining budget, redirects the agent toward a different strategy. This targeted pivoting avoids wasting resources on redundant exploration.

\subsection{Summary}

Budget awareness improves performance through a clear mechanism: it shifts agents from rigid, budget-agnostic behavior to adaptive, resource-aware decision making. Specifically, agents (1) select qualitatively different strategies based on remaining budget, achieving better results with fewer resources; (2) allocate tool calls more deliberately across tool types; and (3) in \method, make principled verification decisions that balance answer confidence against resource availability. The consistent gains across models, benchmarks, and scaling strategies stem from this fundamental behavioral shift---not from simply using more tools, but from using each tool call more wisely.

\section{Case Study}\label{app:case_study}
This section presents outputs from different agents.
To avoid contaminating the internet with the cases shown, we have removed essential details and display only snippets of the agent trajectories.

\subsection{Adaptive Thinking and Planning}\label{app:case_plan}

\begin{figure}[htbp]
    \centering
    \includegraphics[width=0.95\linewidth]{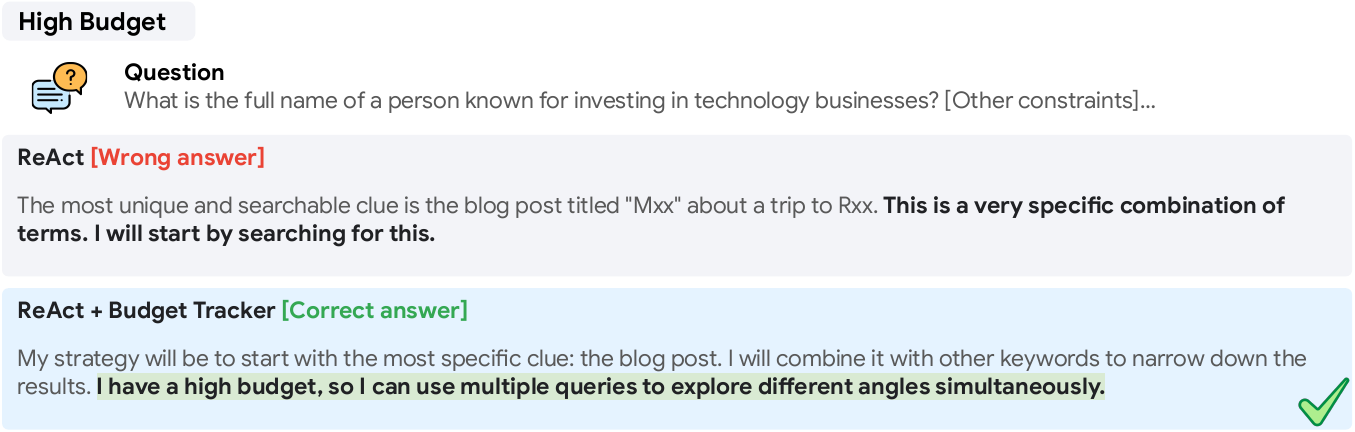}
    \caption{Adaptation to high budget constraints. Being aware of the high budget, the agent chooses to expand its search queries starting from a specific keyword, whereas ReAct immediately applies narrow conditions and fails.}
    \label{fig:case_high_bgt}
\end{figure}

\begin{figure}[htbp]
    \centering
    \includegraphics[width=0.95\linewidth]{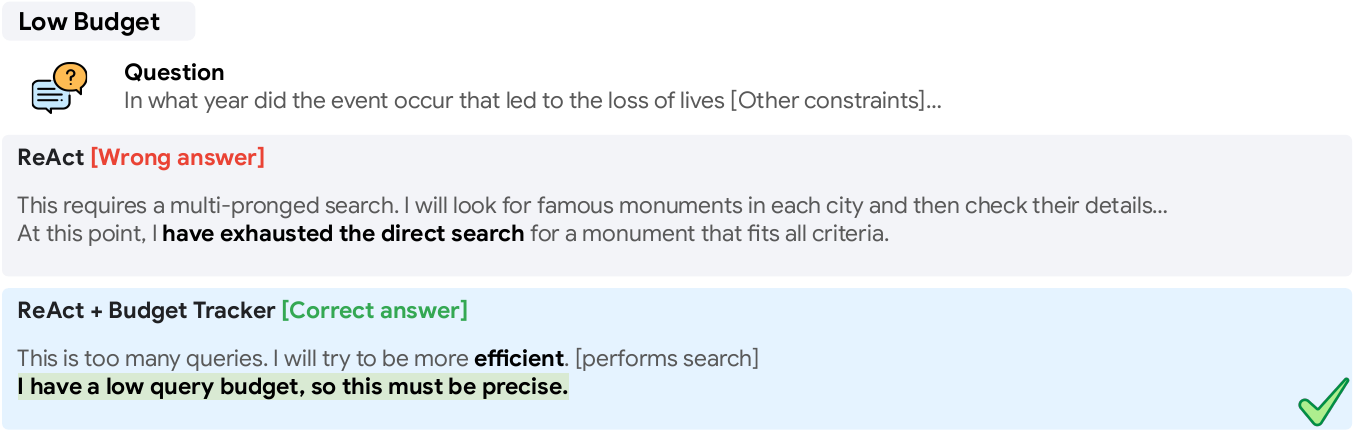}
    \caption{Adaptation to low budget constraints. Under a low budget, ReAct fails due to an exhaustive search strategy. In contrast, budget tracker helps the agent adapt by prioritizing query efficiency and precision, successfully solving the task.}
    \label{fig:case_low_bgt}
\end{figure}

We demonstrate how Budget Tracker enables the agent to dynamically adapt its behavior based on budget awareness, choosing different strategies under high (Figure~\ref{fig:case_high_bgt}) or low (Figure~\ref{fig:case_low_bgt}) budget. These cases are collected from Gemini-2.5-Pro trajectories on BrowseComp. Without budget awareness, ReAct uses a brute-force planning strategy to exhaust direct searches. This causes it to quickly consume all available resources, failing to answer the question correctly.

\subsection{Adaptive Verification}

\begin{figure}[htbp]
    \centering
    \includegraphics[width=0.95\linewidth]{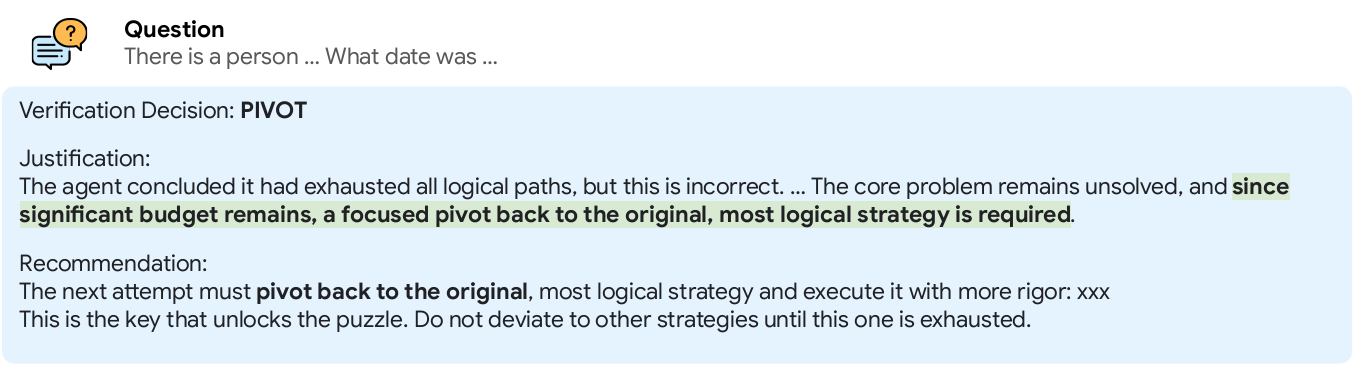}
    \caption{In \method, the verification process dynamically adapts to budget constraints. It successfully identifies the cause of failure and provides strategic recommendations based on the current trajectory. Following this advice allows the agents to better utilize the budget and arrive at the correct answer in the next attempt.}
    \label{fig:case_verification}
\end{figure}

In Figure~\ref{fig:case_verification}, we demonstrate how self-verification module makes budget aware decision based on the trajectory and the resource status. It decides to pivot from the current pass because the remaining budget still allows for further investigation. These cases are collected from Gemini-2.5-Pro trajectories on BrowseComp.

\section{Limitations}

\textbf{More resource constraints.}
While our study presents the first empirical analysis of tool-call budgets, a more realistic and challenging scenario involves managing multiple resource constraints jointly. Examples include token limits, inference latency, and tool-call budgets. Understanding and controlling agent behavior under multi-dimensional constraints is important for deploying scalable systems in real environments.

\textbf{Computational cost.} 
Running tool-augmented agents at scale incurs substantial costs from both API token consumption and tool-call fees, which limits our ability to conduct more extensive repeated experiments or hyperparameter search.

\textbf{Resource allocation.}
Although we observe that budget awareness enables more effective scaling, we do not explore how an agent should allocate its available resources. Our empirical analysis suggests that models often underestimate their actual resource consumption, which can lead to suboptimal performance. Developing accurate resource estimation and principled budget allocation strategies is a promising direction for future work.

\textbf{More intelligent context management.}
Although we adopt several simple context control techniques such as removing tool responses and summarizing intermediate trajectories, more advanced context engineering remains largely unexplored. Designing more effective memory formats and identifying the right balance between context length and performance are important open challenges for building robust and efficient agents.

\end{document}